\documentclass{article}
\newif\ificlrsubmission
\iclrsubmissionfalse
\newcommand{\repourl}{\url{https://github.com/barabbs/IsomorphismNotEnough}}
\PassOptionsToPackage{hyperfootnotes=false}{hyperref}
\usepackage{iclr2027/iclr2027_conference,times}
\ificlrsubmission\else\iclrfinalcopy\fi
\usepackage{hyperref}
\usepackage{url}
\usepackage{latexsym}
\usepackage[T1]{fontenc}    %
\usepackage[utf8]{inputenc} %
\usepackage{microtype}      %
\usepackage{inconsolata}
\usepackage{graphicx}

\usepackage{booktabs}       %
\usepackage{xltabular}      %
\usepackage{amsfonts}       %
\usepackage{nicefrac}       %
\usepackage{enumitem}       %
\usepackage{xcolor}         %
\usepackage[most]{tcolorbox}
\usepackage{multirow}
\usepackage{caption}
\definecolor{termblue}{HTML}{55C8FF}
\definecolor{termpurp}{HTML}{B1B9F9}
\newtcolorbox{terminalbox}{
  colback=black!92,
  coltext=white,
  boxrule=0pt,
  arc=2pt,
  left=6pt,
  right=6pt,
  top=4pt,
  bottom=4pt,
  enhanced,
  fontupper=\ttfamily\scriptsize
}

\definecolor{lean}{HTML}{606060}

\definecolor{GREEN}{HTML}{D9EBD3}
\definecolor{YELLOW}{HTML}{FFF3CC}
\definecolor{ORANGE}{HTML}{FDE5CD}
\definecolor{RED}{HTML}{F4CCCC}
\definecolor{BLUE}{HTML}{C9DBF8}
\definecolor{PURPLE}{HTML}{EBD1DC}
\newrobustcmd{\flagOK}{{\colorbox{GREEN}{\textsc{ok}}}}
\newrobustcmd{\flagISSUE}{{\colorbox{BLUE}{\textsc{issue}}}}
\newrobustcmd{\flagMISMATCH}{{\colorbox{YELLOW}{\textsc{mismatch}}}}
\newrobustcmd{\flagFAIL}{{\colorbox{RED}{\textsc{fail}}}}

\definecolor{TASK}{HTML}{000000}
\newrobustcmd{\specgen}{{\color{TASK}\textsc{spec\_gen}}}
\newrobustcmd{\speciso}{{\color{TASK}\textsc{spec\_iso}}}
\newrobustcmd{\implgen}{{\color{TASK}\textsc{impl\_gen}}}
\newrobustcmd{\proofgen}{{\color{TASK}\textsc{proof\_gen}}}

\newrobustcmd{\lean}[1]{{\color{lean}\texttt{#1}}}
\newcommand{\sorry}{\lean{sorry}}
\newcommand{\genspec}{\lean{generated\_spec}}
\newcommand{\probspec}{\lean{problem\_spec}}

\newcommand{\clever}{\textsc{Clever}}
\newcommand{\humaneval}{\textsc{HumanEval}}
\newcommand{\verina}{\textsc{Verina}}

\input{floats/floatspots}
\putfloat{intro-study}{figure*}{t}{floats/overview}
\putfloat{results-spec-intro}{table}{t!}{tables/results_spec_matrix} %
\putfloat{results-spec-guided}{figure*}{t}{floats/claude_flags_spec} %
\putfloat{results-impl-generated}{table}{t!}{tables/results_impl_matrix} %
\putfloat{results-impl-generated}{figure*}{t}{floats/claude_flags_impl} %
\newcommand{\customindent}{\noindent}

\newcommand\blfootnote[1]{%
  \begingroup
  \renewcommand\thefootnote{}\footnote{#1}%
  \addtocounter{footnote}{-1}%
  \endgroup
}

\title{Isomorphism Is Not Enough: An Agentic Audit of Verified Code Generation in Lean}
\author{Alessandro Sosso, Akhil Arora \& Bas Spitters \\
  Department of Computer Science \\
  Aarhus University \\
  Aarhus, Denmark \\
  \texttt{\{sosso,akhil.arora,spitters\}@cs.au.dk}}

\begin{document}
\maketitle
\lhead{Preprint}

\begin{abstract}
Agentic systems have recently emerged as state-of-the-art approaches for automated theorem proving in formal mathematics. To assess how far these capabilities extend to program verification and how reliably existing benchmarks can measure them, we audit \clever{}, a Lean~4 benchmark for verifiable code generation, using three independent agentic stacks based on Claude Opus~4.6, GPT-5.4, and Kimi K2.5. Our audit finds arguably erroneous reference specifications in 80 of 161 entries, each accepted by the benchmark's authors, distills a taxonomy of benchmark errors, and exposes structural limitations of isomorphism-based scoring, whose signal breaks down when generated and reference specifications formalize different yet defensible readings.
In the process, the stacks generate arguably valid specifications for up to 98.8\% of entries, certify implementations against sound reference specifications for up to 92.3\%, and reach up to 96.2\% end-to-end success over entries with sound premises, suggesting that \clever{} no longer stresses modern agentic provers as much as its design intended.
Looking ahead, we translate our findings into recommendations for future work, with our error analysis doubling as a repair list for \clever{} and a heuristic checklist for the next generation of benchmarks.

\end{abstract}

\ifdefined\blfootnote\blfootnote{Our code, prompts, and per-entry results are available at \repourl{}.}\fi

\section{Introduction}
\label{sec:introduction}
\floatspot{intro-llms}
Large Language Models (LLMs) are known to hallucinate, and even recent ``reasoning'' models may produce arguments that appear coherent but are logically invalid. In contrast, formal logic offers a long-standing and rigorous alternative: mathematical arguments can be expressed in formal languages and mechanically checked for correctness. This approach is embodied in interactive proof assistants based on type theory (e.g., Rocq~\citep{rocq}, Lean~\citep{lean4}, Agda~\citep{norell2007towards}), higher-order logic (e.g., Isabelle~\mbox{\citep{nipkow2002isabellehol}}, HOL4~\mbox{\citep{hol4}}, HOL-Light~\mbox{\citep{harrison1996hollight}}), and set theory (e.g., Mizar~\mbox{\citep{mizar}}, MetaMath~\mbox{\citep{metamath}}).
We focus on the first category, in particular the Lean~4 theorem prover.

\floatspot{intro-progress}
Recent years have seen striking progress in applying LLMs to automated proof construction in type-theoretic settings. Notable systems achieve strong performance on formalized mathematics benchmarks in Lean~4~\citep{hubertOlympiadlevel2025, achimAristotleIMOLevel2025, chenSeedProver2025}, as well as in Isabelle~\citep{lin2024fvel, xu2025isamini} and Rocq~\citep{bayazit2025rocqcasestudy, viennot2025minif2frocq}. These systems employ a range of architectural strategies, including whole-proof generation, proof-state search, iterative self-correction on prover feedback~\citep{linGoedelProverV2ScalingFormal2025, leangTheoremProverJudge2025}, and increasingly, agentic designs that integrate planning, tool use, and iterative refinement.

\floatspot{intro-study}
We study how these advances translate to \emph{verified code generation}, or \emph{vericoding} for short, a setting both practically relevant and structurally challenging. 
It is also far less explored: Lean~4 hosts many benchmarks for formal mathematics but only a handful for verified code generation (Section~\ref{sec:clever_benchmark}), and strong results on the former need not carry over to the latter. Unlike mathematics, program verification requires reasoning about executable artifacts, explicit specifications, and subtle side conditions, all under strict syntactic and semantic constraints. 
To this end, we conduct a systematic \emph{audit} of \clever{}~\citep{thakurCLEVERCuratedBenchmark2025a}, one of the few benchmarks for end-to-end code generation and verification in Lean~4, and of its evaluation methodology. We evaluate three agentic stacks (Claude Code with Opus~4.6, Codex CLI with GPT-5.4, and OpenHands with Kimi K2.5) on its full pipeline, across successive prompt regimes and task variations; the results show what agentic systems can certify, and where isomorphism-based scoring obscures that capability.
Figure~\ref{fig:overview} provides a schematic overview of our experimental pipeline.

\medskip

\floatspot{intro-contributions}
\noindent\underline{Our contributions} are as follows:

\floatspot{intro-audit}
\customindent\textbf{Benchmark audit.}\quad We run the full \clever{} pipeline, in multiple variations of its 4 tasks, with three agentic stacks based on Claude Opus~4.6, GPT-5.4, and Kimi K2.5, on all 161 entries. The audit finds arguably erroneous reference specifications in 80 entries, defects in the benchmark's fixed components that make 4 entries infeasible, and 13 further entries whose NL descriptions admit valid readings that differ from the reference, which isomorphism-based scoring then rejects. The verdicts are stable across prompts, tasks, and stacks, and were reviewed and accepted, entry by entry, by the \clever{} authors. We deliver them as a per-entry annotation and as taxonomies of errors and failure modes, intended as a repair list for \clever{} and a heuristic checklist for future benchmarks.

\floatspot{intro-insight}
\customindent\textbf{Methodological insight.}\quad Our findings expose structural limitations of isomorphism-against-a-ground-truth as an evaluation method for autoformalization: its signal breaks down when generated and reference specifications formalize different yet defensible readings, and it accepts vacuous equivalences between unsatisfiable specifications. They motivate recommendations for the next generation of verification benchmarks: less buggy datasets, more principled evaluation, and more challenging tasks.

\floatspot{intro-saturation}
\customindent\textbf{Saturation.}\quad The three stacks generate arguably valid specifications for 87.0--98.8\% of all entries, certify implementations against 67.9--92.3\% of the sound references, and reach 75.2--96.2\% end-to-end success on sound premises, with the two frontier stacks near the top of each range. Under \clever{}'s own end-to-end criterion they succeed on up to 59 of 161 entries, where the original evaluation reached 1/161.

\floatspot{intro-end}

\section{Methodology}
\label{sec:methodology}

\subsection{Audit scope}
\label{sec:clever_benchmark}
\floatspot{benchmark-clever}
The \clever{} benchmark~\citep{thakurCLEVERCuratedBenchmark2025a} is a curated dataset of 161 problems adapted from \humaneval{}~\citep{chen2021evaluatinglargelanguagemodels}, testing end-to-end automatic code generation and verification in Lean. Each problem is provided as an annotated Lean file containing multiple sections: (1)~the natural language (NL) specification of a function (formatted as a Python-style docstring in a comment), complete with the function signature and example usages; (2)~the Lean~4 signature of the formal specification \genspec{}, with a \sorry{} placeholder to be replaced by the generated specification; (3)~a human-authored ground-truth specification \probspec{} with the same signature; (4)~an \textit{isomorphism} theorem stating the semantical equivalence between \genspec{} and \probspec{}, with placeholder \sorry{}; (5)~the Lean~4 signature of the implementation, with placeholder \sorry{}; and (6)~a \textit{correctness} theorem stating that the implementation satisfies the \probspec{} specification, with placeholder \sorry{}. Helper definitions for problems that require them are also provided, as well as some formal test cases for the implementation.

\floatspot{benchmark-pipeline}
To evaluate the models, the benchmark employs a staged pipeline, defining 4 different tasks grouped in two stages:
\textit{Specification certification} consists of generating the \genspec{} specification (\specgen{}), then proving it is isomorphic to the ground-truth \probspec{} (\speciso{});
\textit{Implementation certification} consists of implementing the function (\implgen{}), then proving it satisfies the \probspec{} specification (\proofgen{}).
The provided testing harness selectively retrieves the sections related to each task; solutions are then evaluated by formatting a new Lean script with the generated elements and testing it for successful compilation and absence of \lean{sorry} keywords.

\floatspot{benchmark-emendation}
\customindent\textbf{Benchmark emendation.}\quad A custom fork of the original repository tracks our fixes in the benchmark and the minor edits to the evaluation scripts required by our setup.
Emendations to the benchmark entries were made in order to fix errors and inconsistencies in formatting, typos and erroneous test cases, as experimentation progressed and issues of this kind emerged. All other elements such as ground-truth specifications, helper functions, and theorem signatures were left untouched.

\floatspot{benchmark-prior-work}
\customindent\textbf{Prior work.}\quad The original paper~\citep{thakurCLEVERCuratedBenchmark2025a} reported an end-to-end ceiling of $1/161$ ($\approx 0.62\%$) under a pass@$k$-seconds protocol with a 600\,s budget, reached by several few-shot and COPRA-based~\citep{thakur2024copra} configurations; the best per-stage figures were $14/161$ ($8.7\%$) for implementation certification and $3/161$ ($1.86\%$) for specification certification. To our knowledge, no later work has exceeded the end-to-end ceiling under the same protocol. Follow-ups report partial pipelines only: Goedel-Code-Prover~\citep{li2026goedelcodeproverhierarchicalproofsearch} reaches $54.0\%$ on the correctness proofs alone, given implementations generated by GPT-5.2 and manually verified, while frontier reasoning models reach at most $23.6\%$; WybeCoder~\citep{gloeckle2026wybecoder} evaluates an imperative translation of \clever{}, not the benchmark itself.
Specification-level property-based testing~\citep{feng2026leetproof} flags 18 entries ($11.2\%$) as defective, 16 of them for underspecified postconditions, and Goedel-Code-Prover's quickcheck finds counterexamples to the correctness theorem, specification and implementation combined, for 10 entries.

\floatspot{benchmark-other-benchmarks}
\customindent\textbf{Other benchmarks.}\quad Most benchmarks for verifiable code generation supply the formal specification and test only the downstream stages: FVAPPS~\citep{dougherty2025fvapps} casts \textsc{APPS} problems as Lean~4 signatures and theorems to implement and prove, the vericoding benchmark~\citep{bursuc2025benchmarkvericodingformally} and AlgoVeri~\citep{zhao2026algoveri} do the same across Dafny, Verus, and Lean, and NTP4VC~\citep{xu2026ntp4vc} isolates the proof stage, discharging verification conditions from existing verification projects. Those that also require a generated specification start from other sources or target other languages: VeriBench~\citep{miranda2026veribench} autoformalizes Python reference programs into Lean~4, VeriContest~\citep{xie2026vericontest} targets Rust with Verus, and VeriEquivBench~\citep{zeng2025veriequivbench} Dafny.

\floatspot{benchmark-vericoding}
\customindent\textbf{Vericoding in Lean.}\quad To our knowledge, at the time of writing only \clever{} and \verina{}~\citep{yeVERINABenchmarkingVerifiable2025} cover the full Lean~4 pipeline from an NL description to a certified implementation, scoring the generated specification against a reference within Lean itself. They differ in how: \verina{} checks soundness and completeness with an LLM-based prover, which rarely settles the check, and otherwise with test suites and property-based testing, whereas \clever{} requires a machine-checked isomorphism proof.

\floatspot{benchmark-preliminary}
\customindent\textbf{Preliminary experiments.}\quad An exploratory study on both benchmarks, restricted to proof generation from the reference specifications and implementations, compared leading agentic and non-agentic provers (Appendix~\ref{app:preliminary-experiments}). Agentic systems based on Claude Code already performed best on the verified-correct portions of both benchmarks, with Aristotle~\citep{achimAristotleIMOLevel2025} close behind, while specialized whole-proof generation models and symbolic tactics trailed substantially. The strongest agents also consistently identified errors in the benchmarks' specifications and implementations, in several cases proposing fixes and proving the corrected statements.
These errors point to the weak spot of \clever{}'s scoring. By accepting a generated specification only through a formal proof against a single reference, it is the stricter check of the two, but also the most exposed to flaws in that reference. The present work therefore audits this mechanism, evaluating agentic systems on the full \clever{} pipeline with its native evaluation infrastructure.

\floatspot{benchmark-end}

\subsection{Experimental setup}
\label{sec:experimental_setup}
\floatspot{setup-stacks}
We evaluated three agentic stacks on all 161 \clever{} entries: the Claude Agent SDK running Claude Opus~4.6, the OpenAI Codex CLI running GPT-5.4, and the OpenHands SDK running the open-weight Kimi K2.5. All three used the same benchmark tasks, prompt regimes, and Lean tooling: \emph{lean-lsp-mcp}~\citep{dresslerOOo0oOoLeanlspmcp2026} exposes via MCP the Lean LSP and provides search tools over the project and Mathlib, while \emph{lean4-skills}~\citep{freerCameronfreerLean4skills2026} injects Lean-specific instructions and commands, tactic best practices and workflow patterns into the model's context.

\floatspot{setup-attempt}
For each attempt, the system initializes a temporary Lean project, creates a file containing the relevant parts of a benchmark entry, and prompts a fresh agent instance to edit it. After completion or timeout, we retain the Lean file, conversation and tool traces, and run metadata. Each agent receives the task description as system prompt: objective, input and output formats, behavior rules, and any guidelines to follow. A short NL request then asks it to fill a specification or implementation (\specgen{}, \implgen{}); theorem-proving tasks (\speciso{}, \proofgen{}) instead invoke the \lean{/lean4:autoprove} command from the \emph{lean4-skills} package, starting a multi-cycle theorem-proving routine. The stacks differ in agent framework, model serving, time budget, and the way skills and commands are delivered; Kimi also ran under an in-run compliance check. Appendix~\ref{app:setup-details} details these differences and the status-flag protocol, summarized below.

\floatspot{setup-prompts}
We successively developed multiple prompt regimes to test whether more structured guidance could overcome the benchmark limitations that emerged. An initial \emph{simple} prompt provided only the task description and a request to fill the relevant section. The \emph{guided} prompt added a short set of instructions for each task, derived by Claude, the highest-performing stack, from an analysis of its own initial run. A further \emph{multiple-specification} prompt was run only with Claude, on entries without a successful guided \speciso{} run, asking it to generate multiple candidate specifications at once.

\floatspot{setup-status}
\customindent\textbf{Status reporting.}\quad After each task, the agent reports \flagOK{} for a completed, compiling solution, \flagFAIL{} for an incomplete attempt, or \flagISSUE{} when it argues that a bug or incorrect assumption makes the task impossible (Appendix~\ref{app:status-flags}). For \speciso{}, \flagMISMATCH{} instead denotes two different but defensible readings of the NL description, with \flagISSUE{} reserved for an erroneous specification (one contradicting the NL description or test cases). For \proofgen{}, the issue may lie in the specification, implementation, or fixed benchmark components. The agents were asked to support these diagnoses with arguments or counterexamples. A handful of flags were amended for technical reasons such as a prose-only answer, with each amendment tracked in the released solution files.

\floatspot{setup-diagnosis}
To extract deeper diagnostic insights, solutions and traces in the \flagISSUE{} category were further analyzed with Claude to identify where the bug originated: in the ground-truth \probspec{}, the generated \genspec{} or both for \speciso{}; and in the specification, implementation, or fixed benchmark components for \proofgen{}. Likewise, \flagMISMATCH{} entries were analyzed to determine whether the generated specification was stronger or weaker than the reference (i.e.~one implied the other), or neither.

\floatspot{setup-cross-checking}
\customindent\textbf{Cross-checking of results.}\quad Reported flags of all tasks directly supported by the benchmark (all \speciso{} runs and \proofgen{} from ground-truth specs) were cross-checked against the benchmark harness: all and only \flagOK{} solutions passed, apart from one correct solution obstructed by a test-case bug and one successful run reported as a timeout. For all other tasks (\specgen{}, \implgen{}, and \proofgen{} from generated specs), \flagOK{} solutions were cross-checked against the Lean typechecker and the absence of \lean{sorry} and \lean{axiom} in the output, with discrepancies in only 3 \proofgen{} runs due to bugs in test cases, all already flagged as \flagISSUE{} at the earlier \implgen{} stage.

\floatspot{setup-annotation}
\customindent\textbf{Merged annotation protocol.}\quad As a sanity check, the solutions and traces of the top-performing stack, Claude Code, were systematically inspected for soundness and underlying structural bugs by the authors. Results of the \speciso{} runs were then merged to produce a single merged annotation for each entry, following fixed rules (no valid specification produced $\to$ \flagFAIL{}; a ground-truth bug in any run $\to$ \flagISSUE{}; otherwise, a successful run $\to$ \flagOK{}; otherwise \flagMISMATCH{}), discussed in Section~\ref{sec:results-spec}.

\floatspot{setup-validation}
\customindent\textbf{Independent validation.}\quad We shared the complete audit with the \clever{} authors: the merged annotation and the per-entry findings of Appendix~\ref{app:complete-findings}, each as a short summary and a long-form description, covering the 80 flagged references, the 4 entries infeasible because of their fixed components, and the 13 \flagMISMATCH{} classifications. They reviewed every finding and accepted all of them; the corrections are being incorporated into the next release of the benchmark. %
This confirmation covers every verdict that defines the sound subsets $C_\mathrm{spec}$, $C_\mathrm{gt}$, and $C_\mathrm{gen}$ used throughout Section~\ref{sec:results}; the validity of generated specifications remains the agents' own judgment, inspected by the authors as described in Section~\ref{sec:results-spec}.

\floatspot{setup-end}

\section{Results}
\label{sec:results}
\floatspot{results-start}
We evaluated the three stacks on all four tasks, with a single attempt per entry and run. This already produced specifications, implementations, and proofs of consistently high quality, often not credited by the benchmark harness, which led us to design the sequence of prompts described below. Appendix~\ref{app:results-details} gives per-entry results.

\subsection{Specification certification}
\label{sec:results-spec}

\floatspot{results-spec-intro}
We score generated specifications in two ways. A specification is \emph{valid} when its \speciso{} run finds it an admissible reading of the NL description (flag \flagOK{}, \flagMISMATCH{}, or \flagISSUE{} with the bug in \probspec{} alone), and \emph{accepted} when its isomorphism to \probspec{} is proved (\flagOK{}, confirmed by the harness), \clever{}'s own criterion. Table~\ref{tab:results-spec-matrix} reports both over all entries and over $C_\mathrm{spec}$, the 80 entries whose \probspec{} and helper definitions the merged annotation found sound.

\floatspot{results-spec-simple}
\customindent\textbf{Simple prompt.}\quad The first minimal prompt gave only the task instructions and the input/output format, without further guidance; as in the original protocol, \specgen{} saw the NL description alone, without example usages or test cases. All stacks produced compiling specifications for every entry at the \specgen{} stage, but only 59, 49, and 34 were \emph{accepted} for Claude, GPT-5.4, and Kimi at the \speciso{} stage. Proof synthesis was rarely the obstacle: besides 4 timeouts, Claude flagged 66 non-isomorphic entries as \flagISSUE{} in the reference, 22 as \flagMISMATCH{}, only 9 as \flagISSUE{} in its own specification, and 1 in both, leaving 147 \emph{valid} specifications. GPT-5.4 and Kimi follow the same pattern, with more failed proofs and own-specification bugs (136 and 121 \emph{valid}). Claude's \flagMISMATCH{} specifications were mostly stronger than the reference (15 against 1, with 6 incomparable), consistent with the references' tendency towards under-restrictive formulations (e.g.~leaving behavior outside the valid input domain unconstrained); Kimi shares this skew (22 against 7), whereas GPT-5.4 reverses it (4 against 12), so the direction also reflects each stack's authoring style.

\floatspot{results-spec-guided}
\customindent\textbf{Guided prompt.}\quad A second prompt sought to align generated specifications with the reference without revealing it. A new Claude instance derived candidate guidelines from the outputs and traces of Claude's own first run, which we refined by hand and stripped of benchmark-specific examples; they covered the structure of the specification, handling of edge cases and degenerate inputs, preference for relational formulations, and Lean~4 syntax. The prompt also fixed a format for the specification body, and the setup now supplied the example usages and formal test cases, aiming to clarify the reading the reference adopted.
Guidance raised the number of \emph{valid} specifications for every stack (to 157, 140, and 123), but not that of \emph{accepted} ones: these rose for GPT-5.4 (49 to 56) and Kimi (34 to 42) and fell for Claude (59 to 56), whose own-specification \flagISSUE{} flags dropped from 10 to 2 while its \flagMISMATCH{} flags rose from 22 to 31. The gap between \emph{valid} and \emph{accepted} persists: on $C_\mathrm{spec}$, Claude's two runs yield a \emph{valid} specification for 77 entries but an \emph{accepted} one for only 61, and the three stacks together reach 65. %

\floatspot{results-spec-multiple}
\customindent\textbf{Multiple specifications prompt.}\quad A \flagMISMATCH{} fails only because the generated specification reads the NL description differently from the reference. To close this gap, a third prompt, run only with Claude, the strongest stack, asked for ``all and only'' the admissible readings of each entry, varying preconditions, edge-case handling, quantifier scope, operation semantics, and property strength, under the same guidelines and test cases. On the 105 entries without a guided \flagOK{}, Claude produced 290 candidates (at most four per entry), of which only 15, spanning 11 entries, proved isomorphic to the reference. Enumerating readings thus recovered few matches: on $C_\mathrm{spec}$, Claude's \emph{accepted} entries rose only from 61 to 65.

\floatspot{results-spec-merged}
\customindent\textbf{Merged annotation.}\quad As a final step, we merged Claude's three runs into a single per-entry annotation and, as a sanity check, manually inspected its outputs and traces for evident errors and inconsistencies. This pass resolved disagreements between runs, caught a lingering structural bug invisible through the flag system (problem~160, where isomorphism succeeded thanks to a flawed helper), and found Claude's classifications and supporting arguments reliable and well-reasoned in all cases; the resulting verdicts on the benchmark were then reviewed and accepted by the \clever{} authors (Section~\ref{sec:experimental_setup}, Independent validation). The annotation counts 81 \flagISSUE{} entries, 80 with an arguably flawed \probspec{} and one (problem~160) with the flawed helper, which form the complement of $C_\mathrm{spec}$; 65 \flagOK{} entries, with at least one \emph{accepted} specification; 13 \flagMISMATCH{} entries, whose sound reference no generated specification ever matched; and 2 \flagFAIL{} entries.

\floatspot{results-spec-end}

\subsection{Implementation certification}
\label{sec:results-impl}

\floatspot{results-impl-intro}
An implementation is \emph{certified} when \proofgen{} proves it correct (\flagOK{}) against the given specification: against the reference, this is \clever{}'s own criterion, scored by the harness; against a generated specification, which the harness cannot natively check, we recompiled every proof independently. A certification holds only relative to its target, so an end-to-end success is \emph{valid} or \emph{accepted} according to its specification. The set of sound entries follows the target: against the reference, $C_\mathrm{gt}$ (78 entries) also excludes entries 32 and 123, whose fixed components make the task infeasible; against generated specifications, $C_\mathrm{gen}$ (157 entries) excludes only the 4 entries with such defects.

\floatspot{results-impl-reference}
\customindent\textbf{Against reference specifications.}\quad Following \clever{}'s protocol, the first run generated implementations from the reference specifications and certified them: Claude, GPT-5.4, and Kimi certified 104, 105, and 77 entries (Table~\ref{tab:results-impl-matrix}). At the \implgen{} stage, Claude's few \flagISSUE{} flags pointed to bugs in the provided test cases, which we then fixed. Most \flagISSUE{} flags blame the specification (42 of Claude's 48), and every such flag, for all three stacks, falls on an entry the merged annotation labels \flagISSUE{}. A flawed reference can reject a correct implementation, while an underspecified one can still be met. On $C_\mathrm{gt}$, the frontier stacks are on par (70 and 72 of 78, 89.7\% and 92.3\%), while Kimi trails (53), mostly through failures and timeouts. Requiring also an \emph{accepted} specification yields \clever{}'s own end-to-end criterion, met by 55, 56, and 35 entries (59 for Claude with the multiple specifications run), against 1 of 161 in the original evaluation, obtained, however, with earlier models, other tooling, and a 600\,s budget.

\floatspot{results-impl-generated}
\customindent\textbf{Against generated specifications.}\quad Two further runs implemented and certified each stack's own simple and guided specifications; at the \implgen{} stage, Claude's three \flagISSUE{} flags in the simple run pointed to test cases inconsistent with the generated specification, which had been written without access to them. Claude certified 144 and 148 entries, GPT-5.4 142 and 142, and Kimi 123 and 119: unlike at the specification stage, guidance brought no consistent gain. The remaining \flagISSUE{} flags stem from generated implementations (e.g.~finite fuel that does not cover all inputs), generated specifications (e.g.~mishandled borderline cases), and occasionally the benchmark's fixed parts (Section~\ref{sec:discussion}). Over both runs, the stacks certify 154, 155, and 140 entries, and at least one stack certifies 158. These proofs, however, hold only relative to each stack's own specification. Requiring that specification to be \emph{valid}, Claude reaches end-to-end success on 151 entries, 96.2\% of $C_\mathrm{gen}$, against 90.4\% for GPT-5.4 and 75.2\% for Kimi; requiring it to be \emph{accepted} lowers the counts to 60, 57, and 38, the same gap between \emph{valid} and \emph{accepted} found at the specification stage.

\floatspot{results-impl-end}

\floatspot{results-end}

\section{Discussion}
\label{sec:discussion}
\subsection{Consistency}
\label{sec:consistency}

\floatspot{consistency-intro}
The audit rests on the claim that a reference specification is flawed, so we first check how consistently this verdict recurs across runs, tasks, and stacks (full analysis in Appendix~\ref{app:consistency}).

\floatspot{consistency-runs}
\customindent\textbf{Across runs.}\quad The verdict is largely stable across prompts: between the simple and guided \speciso{} runs, Cohen's $\kappa$ is 0.73 for Claude, 0.81 for GPT-5.4, and 0.55 for Kimi. Claude flags 73 of the 81 \flagISSUE{} entries in at least two of its three runs; its \flagMISMATCH{} verdicts recur as well, in at least two runs for 12 of the 13 entries.

\floatspot{consistency-tasks}
\customindent\textbf{Across tasks.}\quad Every reference bug raised during \proofgen{} falls on an \flagISSUE{} entry, for all three stacks (44, 41, and 39 flags). The proof stage alone, however, surfaces only about half of the flawed references: on most of the others, the stacks certify an implementation against the flawed specification (34, 33, and 24 entries), more often when the bug weakens it. Specification-level comparison, rather than implementation proof, exposes most reference bugs.

\floatspot{consistency-provers}
\customindent\textbf{Across provers.}\quad GPT-5.4 and Kimi, whose flags the merged annotation does not use, attribute a bug to the reference on 76 and 62 of the 81 \flagISSUE{} entries, and jointly on 77 (95.1\%; $\kappa = 0.85$ against the merged annotation, one-sided Fisher $p = 3.4\cdot10^{-31}$). Their flags on $C_\mathrm{spec}$ fall on 8 entries, all flagged by GPT-5.4 (one also by Kimi) and 5 of them \flagMISMATCH{} entries, suggesting that GPT-5.4 is quick to call a defensible reading a bug. No stack ever proves a flawed reference valid: the few isomorphisms accepted on \flagISSUE{} entries reproduce the same bug, reuse the same flawed helper, or are vacuous, with both specifications unsatisfiable on some input. Capability differences surface mainly as failures (on 27, 59, and 170 of 805 certification attempts) and as \flagMISMATCH{} readings in place of bug claims, and only rarely as contradictions. The \flagISSUE{}/\flagMISMATCH{} boundary is where the stacks diverge most: only 8 of the 13 \flagMISMATCH{} entries receive a \flagMISMATCH{} reading from all three.

\floatspot{consistency-end}

\subsection{Taxonomy}
\label{sec:taxonomy}

\floatspot{taxonomy-intro}
We now turn from how often the stacks succeed to why they fail. For this analysis we restrict ourselves to Claude, the strongest-performing stack. We had Claude classify the cause behind each \flagISSUE{} and \flagMISMATCH{} flag of its runs, and grouped the diagnoses by the component at fault~--- ground-truth specifications, generated specifications, implementations, and the benchmark's fixed parts~--- and, for \flagMISMATCH{} entries, by the authoring choice that separates the two readings. The resulting taxonomies (Appendix~\ref{app:taxonomy}) serve both as a diagnosis of \clever{} and as a heuristic checklist of pitfalls for the authors of future benchmarks.

\floatspot{taxonomy-components}
A faithful specification must have the right strength and commit to one reading of an often ambiguous NL description: it must formalize the \emph{intent} behind the prose~\citep{lahiriIntentFormalizationGrand2026}. The taxonomies show how each component falls short of this goal:
\begin{itemize}[leftmargin=*,itemsep=1pt,topsep=2pt]
  \item \textbf{Ground-truth specifications} fail mostly through Lean-encoding pitfalls (48 of the 80 flawed references), led by \emph{conjunct-as-guard} patterns, $(P \wedge Q)$ in place of $(P \to Q)$ (15 problems), and \emph{precedence bugs} (12); the other errors are semantic, most often a wrong interpretation of the NL description (21).
  \item \textbf{Claude's generated specifications} show the inverse profile: their bugs are almost only semantic, mostly missing constraints. Guidance cut them from 10 to 2 per run, but they resurfaced when the multiple specifications prompt pushed Claude to explore the space of readings.
  \item \textbf{Mismatches} reflect authoring choices rather than errors, most often on the boundary of the input domain (73 of 142 assignments).
  \item \textbf{Implementations} were rarely at fault (7 cases in 5 problems), for instance through bounded fuel that does not cover all inputs.
  \item \textbf{Fixed benchmark components} make four entries infeasible: two hinge on open conjectures (problems 123 and 39, on the Collatz sequence and on Fibonacci primes), one ships a faulty helper (160), and one lacks a precondition (32, Appendix~\ref{app:notable-examples}).
\end{itemize}

\floatspot{taxonomy-end}

\subsection{Resources and contamination}

\floatspot{resources-intro}
Appendix~\ref{app:resources} reports statistics on token usage, runtimes, and costs. Appendix~\ref{app:contamination} discusses training-data contamination: why the audit's findings do not depend on it, and an exploratory control with GPT-5.1, a model trained before \clever{} became public.

\floatspot{resources-end}

\subsection{Recommendations}
\label{sec:recommendations}

\floatspot{recommendations-intro}
The same findings also expose structural limitations in how \clever{} is designed and evaluated, and suggest three axes along which the next generation of vericoding benchmarks should improve.

\floatspot{recommendations-datasets}
\customindent\textbf{Less buggy datasets.}\quad The error analysis of Section~\ref{sec:taxonomy} and Appendix~\ref{app:taxonomy} doubles as a repair list for \clever{} and an authoring checklist for new datasets. Of the 80 bugged ground-truth specifications, 48 exhibit recurring Lean-encoding pitfalls of exactly the kind an authoring-time ``linting'' pass (e.g.~property-based testing plus an agentic judge) is designed to catch.

\floatspot{recommendations-evaluation}
\customindent\textbf{More principled evaluation.}\quad For isomor\-phism-based scoring to remain a meaningful signal, the generated and reference specifications must come from a shared intent: more precise NL descriptions, explicit formalization hints (e.g.\ pre-committing to a type signature), or an iterative disambiguation pass in which an LLM judge surfaces and resolves multiple admissible readings until none remains. Scoring must also rule out vacuous successes: \clever{}'s isomorphism theorem, of the form $\forall f.\,(\forall x.\,P(f,x)) \leftrightarrow (\forall x.\,Q(f,x))$, holds for any two specifications that no implementation can satisfy, as in two isomorphisms Claude proved on the flawed entries~67 and~160 (Appendix~\ref{app:consistency-provers}); a pointwise statement, or a witness implementation certified against each specification, would close this gap. Where formal isomorphism is intractable, the reviewed diagnoses and the attribution differences across stacks suggest combining evaluators, LLM-based and deterministic: agentic judges cross-checked by property-based testing of the specifications themselves, and end-to-end synthesis against the model's own specification, which surfaced vacuous, contradictory, or under-constrained specifications that the equivalence check had missed.

\floatspot{recommendations-tasks}
\customindent\textbf{More challenging tasks.}\quad Once the evaluation is trustworthy, benchmarks should turn to harder sources and longer horizons~--- multi-function or repository-level verification, richer data structures and side conditions~--- beyond \humaneval{}-style single functions.

\floatspot{recommendations-closing}
These changes would restore the reliability of the evaluation signal; making it discriminative again for modern agentic provers is the task these three axes leave open.

\floatspot{recommendations-end}

\section{Conclusions}
\label{sec:conclusions}
\floatspot{conclusions-audit}
We audited \clever{}, one of the few benchmarks that score the full pipeline from an NL description to a certified Lean~4 implementation, with three independent agentic stacks on all 161 entries. Under the benchmark's own scoring, the stacks appear far from solving it: no stack passes the isomorphism test on more than 67 entries, or end to end against the reference on more than 59. Under inspection, most of the gap belongs to the benchmark: 80 reference specifications are arguably erroneous, 4 entries are infeasible because of their fixed components, and 13 more admit valid readings the reference does not share. These verdicts recur across prompts, tasks, and stacks, and were independently confirmed by the benchmark's authors. On sound premises, the stacks certify implementations for up to 92.3\% of entries and reach up to 96.2\% end to end, with the two frontier stacks within two entries of each other on implementation certification: the benchmark no longer separates them, and much of its remaining difficulty is an artifact of its evaluation.

\floatspot{conclusions-lesson}
The structural lesson concerns isomorphism against a single reference as a scoring method for autoformalization. Its signal presumes that the reference formalizes the only defensible reading of the prose, which our mismatch entries show it does not always do; the gap between valid and accepted specifications persisted for every stack under guidance, and for Claude even when enumerating readings; and its theorem form accepts vacuous equivalences between unsatisfiable specifications. Future benchmarks should catch the Lean-encoding pitfalls of our taxonomy at authoring time; restore a shared intent before scoring, or combine evaluators where isomorphism cannot decide; and, once the signal is trustworthy, raise the difficulty beyond single functions. Our per-entry annotation and taxonomies are released as a repair list for \clever{} and a checklist for that next generation.

\floatspot{conclusions-end}

\subsection*{Reproducibility statement}
Our repository (\repourl{}) releases the agent harnesses and prompts, our fork of \clever{} (Section~\ref{sec:clever_benchmark}), the per-entry run records and findings, and the scripts behind every figure. To avoid contaminating \clever{}, the public records replace the generated specifications, implementations, and proofs, and the agent traces, with SHA-256 commitments; the complete records are available from the authors on request and can be checked against these commitments. Section~\ref{sec:experimental_setup} and Appendix~\ref{app:setup-details} describe the configurations and pinned tool versions; Appendices~\ref{app:results-details}--\ref{app:resources} and~\ref{app:complete-findings} report per-entry outcomes and further statistics. Since the models are served through endpoints that may change, re-runs may differ on individual entries, but every reported outcome can be re-verified with Lean from the complete records.

\ificlrsubmission\else
\subsubsection*{Acknowledgments}
We thank the Ethereum foundation for partially funding this research.
Arora’s lab is partly supported by grants from the Novo Nordisk Foundation (NNF24OC0099109), the Pioneer Centre for AI, and EU Horizon 2020 (101168951). We also gratefully acknowledge generous gifts from Microsoft and It-vest - networking universities.
The authors acknowledge the AI Lab at the Faculty of Natural Sciences, Aarhus University, for providing AI infrastructure, platforms, models, services, and technical support used in this work.\footnote{Further information is available at https://nat.au.dk/ailab.}

\fi

\bibliographystyle{iclr2027/iclr2027_conference}
\bibliography{bibliography}

\clearpage
\appendix

\section{Preliminary Experiments}
\label{app:preliminary-experiments}
\begin{figure}[b!]
  \centering
  \includegraphics[width=0.9\textwidth]{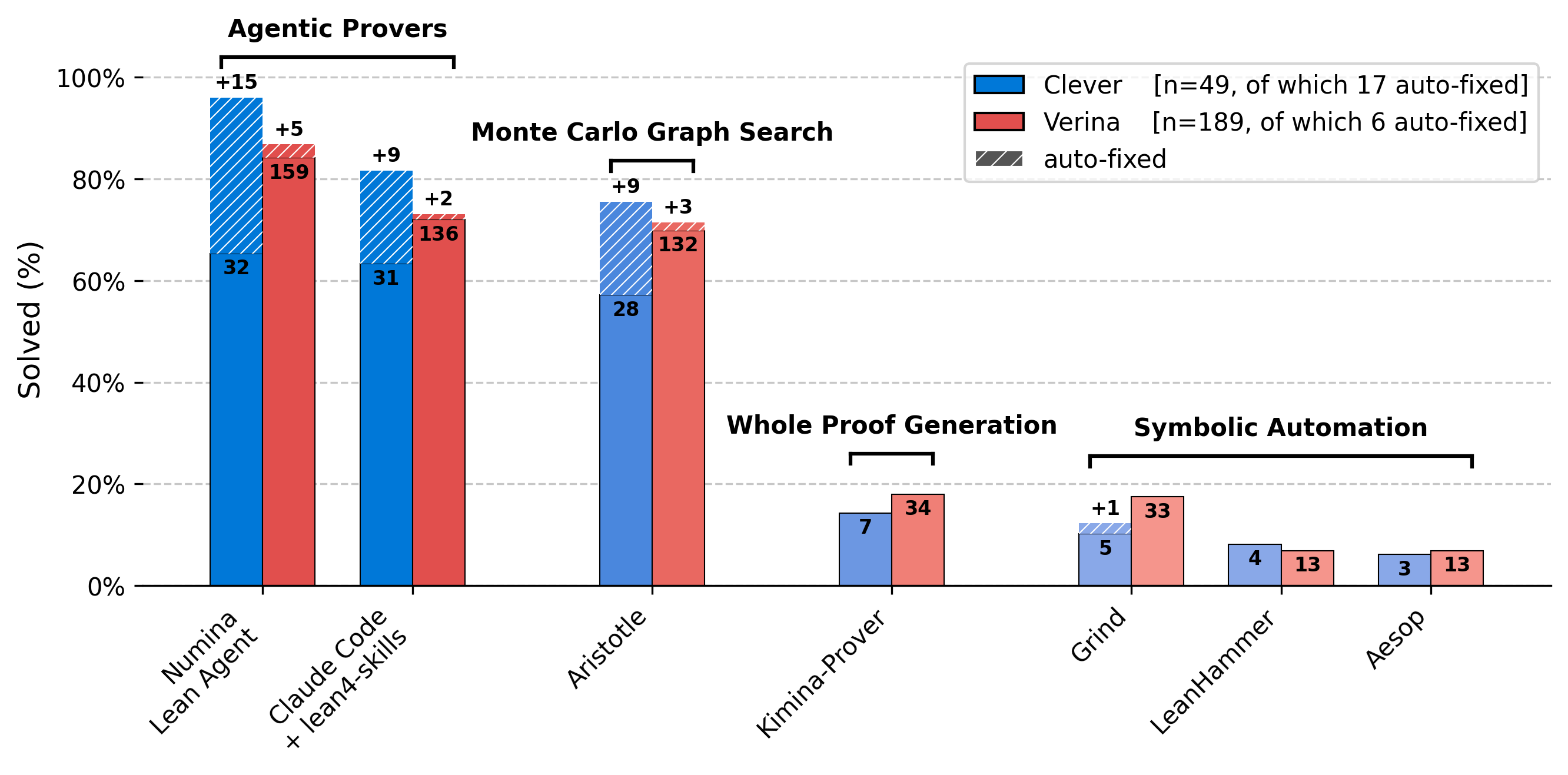}
  \vspace{-0.25cm}
  \caption{Overview of prover performance on the \clever{} and \verina{} benchmarks. Solid bars indicate the number of tests solved on the verified-correct portions of each dataset, while the patterned bar segments above show additional solved instances obtained after correcting erroneous benchmark entries. Numbers annotated with a ‘+’ denote the corresponding count of these additional solves.}
  \label{fig:prelim_experiments_overview}
\end{figure}

In prior experiments, the authors evaluated a suite of automated provers (agentic, specialized, and symbolic) on proof generation on the \clever{}~\citep{thakurCLEVERCuratedBenchmark2025a} and \verina{}~\citep{yeVERINABenchmarkingVerifiable2025} datasets, finding that, in that setup, agentic systems substantially outperformed the alternatives. We summarise its findings here, both because it directly motivates the present work and because some of its observations reappear, refined, in our current results.

\paragraph{Setup.}
The earlier study used a different experimental setup than the one of Section~\ref{sec:experimental_setup}. It focused exclusively on proof generation starting from ground-truth specifications and implementations, a task outside the scope of either benchmark's provided harness (in Figure~\ref{fig:overview}, the setup illustrated in the dashed portion of the diagram in the top-right corner). For this reason it relied on a custom distillation of the two datasets rather than on their provided evaluation infrastructure: derivative versions (denoted \clever{}-P, \verina{}-P) were obtained by reusing the ground-truth specifications and implementations, %
with other variants distilled to meet requirements for specific provers; NL descriptions were not provided. This restriction also limited the \clever{} subset to the 43 entries (out of 161) for which an implementation was available, plus the benchmark's 6 example entries. The agentic configurations were further built on Claude Opus~4.5 and on an earlier release of \emph{lean4-skills}~\citep{freerCameronfreerLean4skills2026} that predated the current \lean{/autoprove} command, relying instead on the \lean{/fill-sorry} command, and with three independent one-shot attempts per problem. The benchmark releases evaluated were \clever{} v1.4.0 and a contemporary \verina{} build, both of which have since been partially patched in light of the bugs reported by those experiments.

\paragraph{Provers.}
The evaluation covered three families. The \emph{agentic} family included Claude Code with the earlier version of \emph{lean4-skills} and \emph{lean-lsp-mcp}~\citep{dresslerOOo0oOoLeanlspmcp2026}, and the \emph{Numina-Lean-Agent}~\citep{liuNuminaLeanAgentOpenGeneral2026,ProjectnuminaNuminaleanagent2026}, a wrapper around Claude Code combining it with a fork of \emph{lean-lsp-mcp} extended with semantic search, an informal-prover and a discussion-partner tool, run with \texttt{max\_rounds=3}. The \emph{whole-proof generation} family~--- specialized models trained or fine-tuned for formal verification, that interleave chain-of-thought reasoning with formal code generation~--- was represented by Kimina-Prover-72B~\citep{wangKiminaProverPreviewLarge2025}, the highest-performing open model of its kind on \textsc{MiniF2F}, with DeepSeek-Prover-V2~\citep{renDeepSeekProverV2AdvancingFormal2025} and Goedel-Prover-V2~\citep{linGoedelProverV2ScalingFormal2025} in the same class. The \emph{symbolic} family covered Grind~\citep{lean_grind_manual}, LeanHammer~\citep{joshcluneJOSHCLUNELeanHammer2026}, Aesop~\citep{limpergAesopWhiteBoxBestFirst2023}, and Canonical~\citep{normanCanonicalAutomatedTheorem2025}. Aristotle~\citep{achimAristotleIMOLevel2025}, Harmonic's high-level proof-state-aware Monte Carlo Graph Search system, was tested alongside as a separate state-of-the-art baseline (most notably achieving gold-medal performance at IMO 2025).

\paragraph{Results.}

The evaluation revealed a clear performance hierarchy, as displayed by Figure~\ref{fig:prelim_experiments_overview}: agentic systems based on Claude Code performed best, with Aristotle trailing by a small margin; Kimina-Prover displayed pass rates much lower than reported on \textsc{MiniF2F} (7/32 on \clever{}-P, 34/183 on \verina{}-P), while symbolic provers provided a stable but significantly lower baseline, consistent with their design as gap-fillers between assertions rather than full proof generators. Considering only the verified-correct portions of the datasets, Numina-Lean-Agent solved all 32 \clever{}-P entries, while Claude Code with \emph{lean4-skills} and Aristotle solved 31 and 28 respectively; out of the 183 correct \verina{}-P entries, the same three provers solved 159, 136 and 132. The high performance of the agentic systems was attributed largely to their direct access to compiler feedback through their search and tool-use loop, the main documented difference in implementation between the various approaches; conversely, the whole-proof-generation models were tested in a single one-shot attempt without successive refinement on compiler feedback, and their performance might improve substantially under iterative protocols.

\begin{table}[h]
  \newcommand{\notavailable}{\raisebox{0.3ex}{\textcolor{gray}{\tiny N/A}}}

  \caption{Number of solved tests by each prover over the correct entries in the tested benchmarks. Numbers in brackets represent the number of solved entries among the auto-fixed ones. %
  }
  \centering
  \footnotesize
  \begin{tabular}{lllrlrllrlrl}
                               & & \multicolumn{4}{c}{\clever}                                             &                      & \multicolumn{4}{c}{\verina}                               \\ \cline{3-6} \cline{8-11}
    \textbf{Prover}            & & \multicolumn{2}{c}{examples} & \multicolumn{2}{c}{human\_eval} & \multicolumn{1}{c}{} & \multicolumn{2}{c}{basic} & \multicolumn{2}{r}{advanced} \\ \hline
    Numina Lean Agent          & & 5                & (+1)              & 27            & (+14)           &                      & 102         & (+2)          & 57           & (+3)          \\
    Claude Code + lean4-skills & & 5                & (+1)              & 26            & (+8)            &                      & 96          &               & 40           & (+2)          \\
    Aristotle                  & & 5                &                   & 23            & (+9)            &                      & 89          & (+1)          & 43           & (+2)          \\
    Kimina-Prover              & & 4                & \notavailable     & 3             & \notavailable   & \multicolumn{1}{c}{} & 31          & \notavailable & 3            & \notavailable \\
    Grind                      & & 2                &                   & 3             & (+1)            &                      & 31          &               & 2            &               \\
    LeanHammer                 & & 1                &                   & 3             &                 &                      & 13          &               & 0            &               \\
    Aesop                      & & 1                &                   & 2             &                 &                      & 13          &               & 0            &               \\
    Canonical                  & & 0                &                   & 0             &                 &                      & 0           &               & 0            &               \\ \hline
    \textit{number of entries} &                    & & 5                & (+1)              & 27            & (+16)           &                      & 106         & (+2)          & 77           & (+4)
  \end{tabular}
  \label{tab:results-passrates}
\end{table}

A second, less expected finding was that Claude Code and Aristotle consistently identified errors in the benchmarks themselves~--- Aristotle through its goal-negation search mechanism, Claude Code through its NL replies (sometimes already proposing fixes and certifying the fixed statements without being prompted to do so). Cumulatively, the provers identified 17 erroneous entries in \clever{} and 6 in \verina{}. When subsequently given the NL descriptions and the previous proof attempts, Claude Code provided fixes for all of them and proved a substantial fraction of the corrected statements: Numina-Lean-Agent proved 15/17 and 5/6 of the fixed \clever{} and \verina{} statements respectively, while Claude Code with \emph{lean4-skills} proved 9/17 and 2/6, and Aristotle 9/17 and 3/6. Most of the fixes addressed wrong specification logic (9 cases), implementation bugs (6) and missing preconditions (5), with a few off-by-one and boundary errors (3).

\begin{table}[h]
  \caption{Number of fixes per benchmark by kind.}
  \centering
  \footnotesize
  \begin{tabular}{llclclc}
    & & \clever{} & & \verina{}                 & & Total   \\ \hline
    {Wrong Specification Logic}  & & 7      &                      & {2} & & 9  \\
    {Implementation Bugs}        & & 4      &                      & {2} & & 6  \\
    {Missing Preconditions}      & & 3      &                      & {2} & & 5  \\
    {Off-by-One/Boundary Errors} & & 3      &                      & {}  & & 3  \\ \hline
    {Total}                      & & 17     &                      & {6} & & 23
  \end{tabular}
  \label{tab:fixes-summary}
\end{table}

\section{Setup Details}
\label{app:setup-details}
This appendix complements Section~\ref{sec:experimental_setup} with the status-flag protocol and the differences between the three agentic stacks.

\subsection{Common protocol}
All stacks work on the same Lean project, pinned to Lean~v4.24.0 and the matching Mathlib release, with the same \emph{lean-lsp-mcp} server and the same commit of \emph{lean4-skills}. Each attempt starts a fresh agent on a fresh copy of the project; each entry receives a single attempt per task (pass@1).

\subsection{Status flags}
\label{app:status-flags}

At the end of each task, the agent is asked to reply with a single status flag. The base prompt, offers three flags:
\begin{itemize}[leftmargin=*,itemsep=1pt,topsep=2pt]
  \item \flagOK{}: task completed, no \sorry{} remains, code compiles;
  \item \flagFAIL{}: unable to complete, \sorry{} remains or code does not compile;
  \item \flagISSUE{}: task is impossible due to bugs or incorrect assumptions, and rigorous arguments in support have been provided.
\end{itemize}
For \speciso{}, the prompt splits \flagISSUE{} in two:
\begin{itemize}[leftmargin=*,itemsep=1pt,topsep=2pt]
  \item \flagMISMATCH{}: equivalence is unprovable because the specifications differ, yet both are valid interpretations of the NL description and test cases; rigorous arguments in support have been provided;
  \item \flagISSUE{}: equivalence is unprovable because at least one specification is erroneous (contradicts the NL description or test cases); rigorous arguments in support have been provided.
\end{itemize}
In the multiple specifications run, the agent reports one such flag per candidate file.

In later \proofgen{} runs, a prompt that offered \flagMISMATCH{} was used as well, although for homogeneity we merged these flags into \flagISSUE{}. %
This is consistent with the flags' meaning, since in implementation tasks the specification is provided, so an implementation that cannot meet it is an error, not an alternative reading. All changes are tracked in the released records.

\subsection{Differences between stacks}
\label{app:stack-differences}

\begin{table}[h]
  \caption{Structural differences between the three agentic stacks. All share the benchmark, prompts, Lean project, MCP server, and \emph{lean4-skills} commit.}
  \label{tab:stack-differences}
  \centering
  \small
  \begin{tabular}{@{}l p{0.2\textwidth} p{0.25\textwidth} p{0.25\textwidth}@{}}
    \toprule
    & Claude & GPT-5.4 & Kimi \\
    \midrule
    Agent framework & Claude Agent SDK & Codex CLI & OpenHands SDK \\
    Model serving & vendor & vendor & open weights, local cluster \\
    Skills & native plugin & native skill directory & native plugin \\
    \lean{/lean4:autoprove} & native command & expanded into the prompt & keyword trigger \\
    Time budget & 1\,h & 1\,h & 4\,h \\
    Context management & framework default & framework default & condenser at 99\% of the context window \\
    \bottomrule
  \end{tabular}
\end{table}

\paragraph{Commands in Codex.}
The Codex CLI has no notion of user-defined commands. For the proof tasks, we therefore expand \lean{/lean4:autoprove} into its full instructions and embed them in the prompt, while the skills themselves are exposed natively.

\paragraph{Compliance check for Kimi.}
In early runs, Kimi often edited parts of the Lean file it was not allowed to change, for instance adding a hypothesis to the theorem it had to prove or rewriting a fixed specification, and sometimes dropped the tags that delimit the file's sections, which the evaluation harness then scores as \sorry{}. Some of these edits turned failed attempts into reported successes. The OpenHands harness therefore validates the file after the agent's main turn and before asking for the status flag:\emph{Check A (tags)}: every section tag of the original file is still present; \emph{Check B (fixed content)}: the non-editable sections and the preamble are unchanged, no axiom was added, and no fixed signature was redeclared; \emph{Check C (flag)}: the status reply is exactly one of the allowed flags.

On a violation of Check~B, the harness restores the original text itself; on a violation of Check~A, it asks the agent to restore the tags. In both cases a corrective prompt explains the violation, and the agent continues its attempt. Each check allows up to three retries; exhausting them records \flagFAIL{} without a status turn. Check~C re-asks for the flag up to three times, again falling back to \flagFAIL{}.

\clearpage
\section{Detailed Breakdown of Results}
\label{app:results-details}
\begin{figure}[h!]
  \centering
  \includegraphics[width=\textwidth]{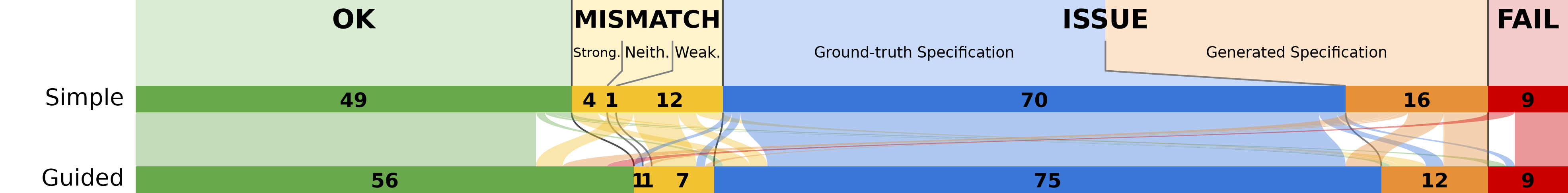}

  \vspace{0.5em}
  \includegraphics[width=\textwidth]{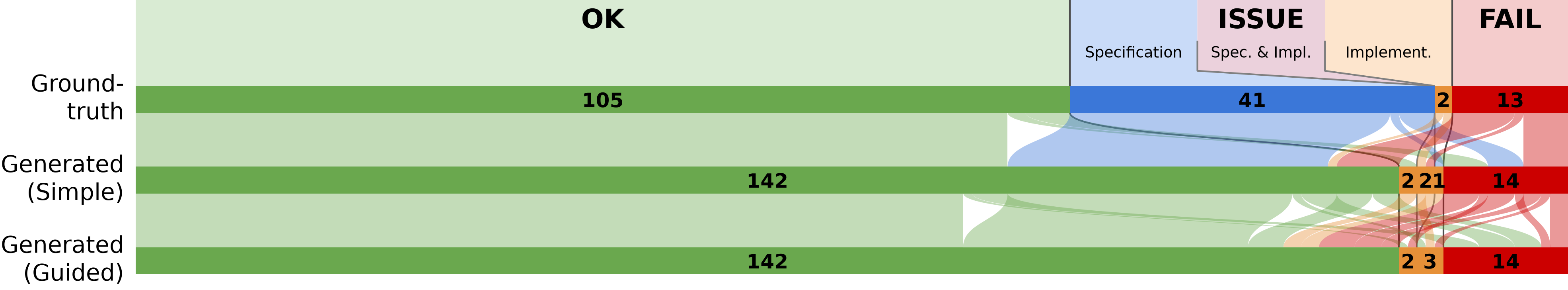}
  \caption{Distribution of GPT-5.4's output flags for the simple and guided \speciso{} runs (top) and for \proofgen{} against the reference and against its simple and guided generated specifications (bottom); bands follow each entry from one run to the next. Colors as in Figure~\ref{fig:claude-flags}; \flagFAIL{} includes timeouts. GPT-5.4 did not take part in the multiple specifications run, and its \proofgen{} \flagMISMATCH{} flags are merged into \flagISSUE{} (Appendix~\ref{app:status-flags}).}
  \label{fig:codex-flags}
\end{figure}

\begin{figure}[h!]
  \centering
  \includegraphics[width=\textwidth]{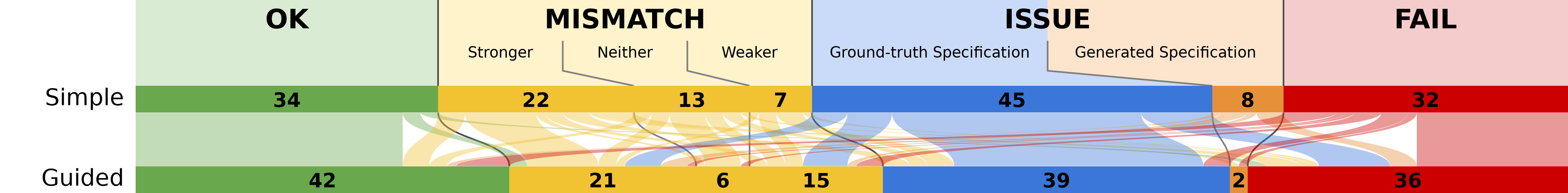}

  \vspace{0.5em}
  \includegraphics[width=\textwidth]{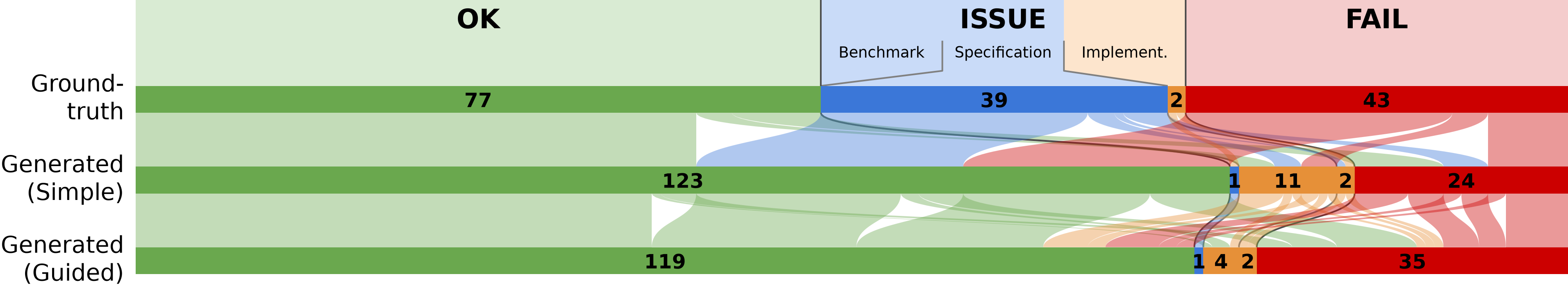}
  \caption{Distribution of Kimi's output flags for the simple and guided \speciso{} runs (top) and for \proofgen{} against the reference and against its simple and guided generated specifications (bottom); bands follow each entry from one run to the next. Colors as in Figure~\ref{fig:claude-flags}; \flagFAIL{} includes timeouts. Kimi did not take part in the multiple specifications run, and its \proofgen{} \flagMISMATCH{} flags are merged into \flagISSUE{} (Appendix~\ref{app:status-flags}).}
  \label{fig:kimi-flags}
\end{figure}

\begin{figure}[h!]
  \centering
  \includegraphics[scale=0.73]{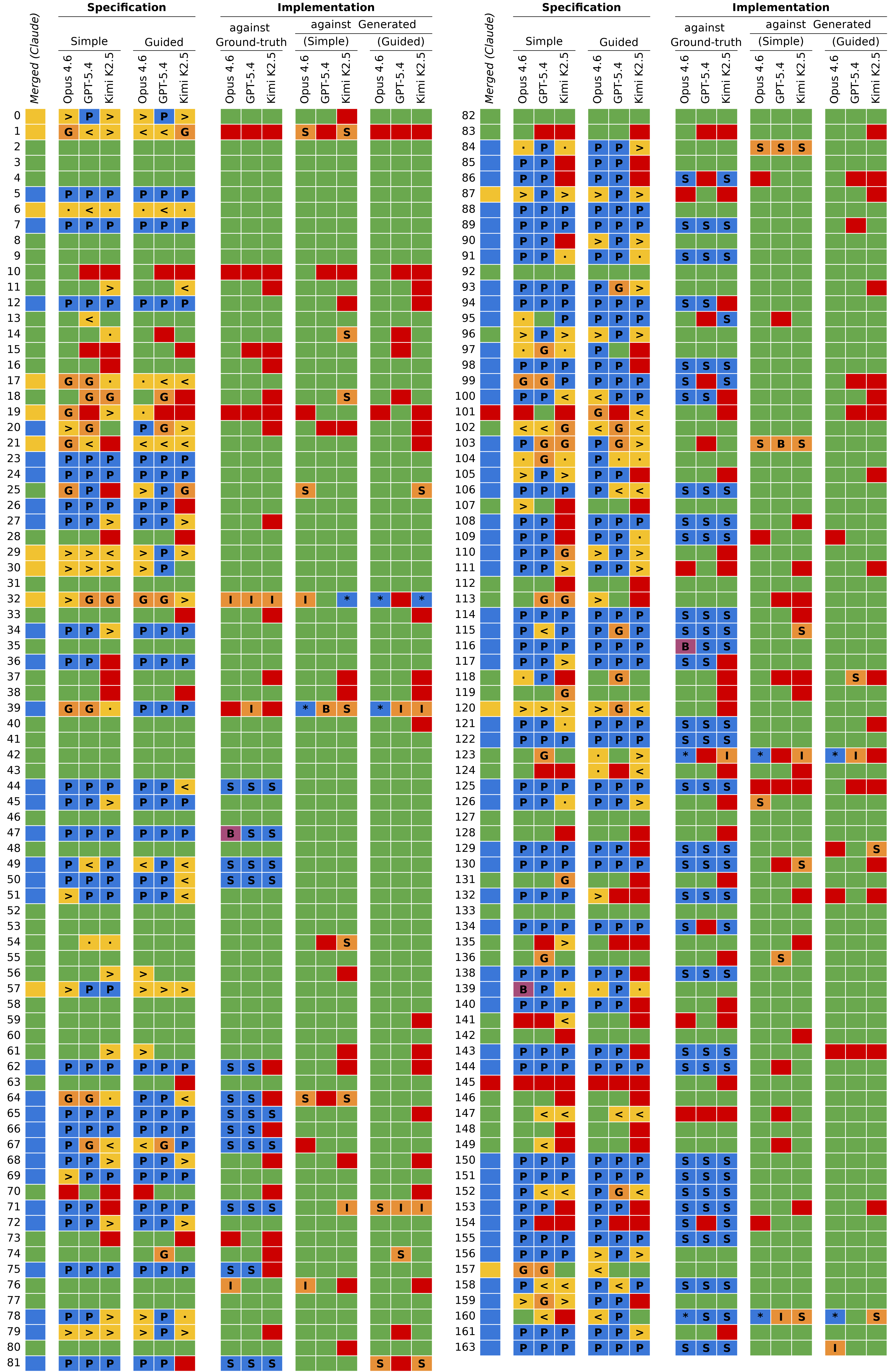}
  \caption{Per-entry flags of the three stacks over all 161 \clever{} entries: \speciso{} under the simple and guided prompts, and \proofgen{} against the reference and against each stack's simple and guided generated specifications. The leftmost column is Claude's merged annotation. Colors as in Figure~\ref{fig:claude-flags}; red includes timeouts. Letters give the source of an \flagISSUE{}: P, G, or B for the reference, the generated, or both specifications in \speciso{}; S, I, or B for the specification, the implementation, or both in \proofgen{}; an asterisk marks a defect in fixed benchmark components. Symbols give the direction of a \flagMISMATCH{}: $>$ for a generated specification stronger than the reference, $<$ for weaker, $\cdot$ for neither.}
  \label{fig:three-stack-grid}
\end{figure}

\begin{figure}[h!]
  \centering
  \includegraphics[scale=0.73]{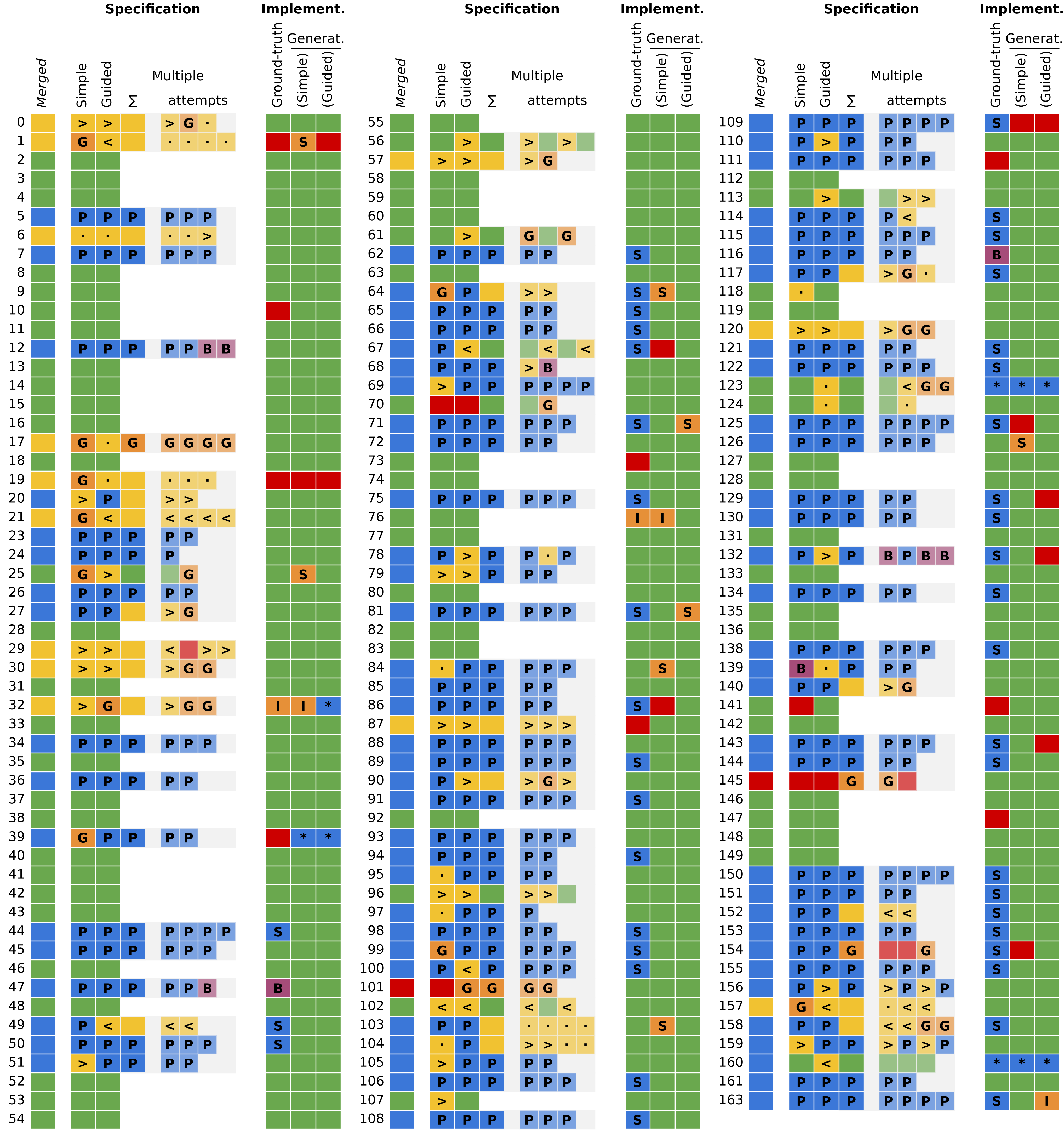}
  \caption{Claude's per-entry outcomes. Specification columns: \speciso{} under the simple and guided prompts, and the multiple specifications run, both aggregated per entry ($\Sigma$) and per candidate (lighter shades); entries already \flagOK{} in an earlier run were not re-tested and are left blank. Implementation columns: \proofgen{} against the reference and against the simple and guided generated specifications. The leftmost column is the merged annotation. Colors, letters, and symbols as in Figure~\ref{fig:three-stack-grid}.}
  \label{fig:claude-results-grid}
\end{figure}

This appendix collects the per-entry outcomes behind Section~\ref{sec:results}.
Figures~\ref{fig:codex-flags} and~\ref{fig:kimi-flags} show how the flags of each run of GPT-5.4 and Kimi are distributed, in the same format as Claude's Figure~\ref{fig:claude-flags}.
Figure~\ref{fig:three-stack-grid} lists every run-level flag of the three stacks for all 161 entries, next to Claude's merged annotation, so that agreements and disagreements between stacks can be read entry by entry; Appendix~\ref{app:consistency} quantifies them.
Figure~\ref{fig:claude-results-grid} details Claude's outcomes, including each candidate of the multiple specifications run.
In both grids, letters give the diagnosed source of an \flagISSUE{} (in \speciso{}, P, G, B for the reference, generated, or both specifications; in \proofgen{}, S, I, B for the specification, the implementation, or both; $*$ for the fixed benchmark components), and symbols the direction of a \flagMISMATCH{} (\texttt{>}, \texttt{<}, $\cdot$ for a generated specification stronger than, weaker than, or incomparable with the reference).

\clearpage

\clearpage
\section{Consistency Analysis}
\label{app:consistency}
This appendix checks how consistently the audit's central verdict, that a reference specification \probspec{} is flawed, recurs across prompt runs, across tasks, and across the three stacks. We call \emph{reference-bug flag} an \flagISSUE{} attributed to \probspec{} (sources \texttt{PROB} or \texttt{BOTH}) in a \speciso{} run, or to the specification (\texttt{SPEC} or \texttt{BOTH}) in \proofgen{} against the reference; proofs against generated specifications never count. The \emph{flawed entries} are the 81 entries annotated \flagISSUE{} in the merged annotation (Section~\ref{sec:results-spec}); the other 80 form $C_\mathrm{spec}$. We write S and G for the simple and guided \speciso{} runs, M for Claude's multiple specifications run, where an entry counts as flagged if any of its candidates is, and PG for \proofgen{} against the reference.

Since the merged annotation derives from Claude's runs, Claude's agreement with it is not independent evidence; we report it only where it describes the stability of those runs. GPT-5.4 and Kimi share prompts, benchmark, and Lean tooling with Claude, but neither model nor agent framework, and their flags do not contribute to the merged annotation. Agreement is measured with unweighted Cohen's $\kappa$, association with a one-sided exact Fisher test, and uncertainty with exact 95\% Clopper--Pearson intervals; all $p$-values are nominal, without correction for multiple testing. %

\subsection{Across runs}

\begin{table}[h]
  \caption{Reference-bug flags in the simple (S) and guided (G) \speciso{} runs, over all 161 entries. $+$/$-$: flagged or not in each run. Claude's either-run counts are omitted, since the annotation derives from them.}
  \label{tab:consistency-runs}
  \centering
  \small
  \begin{tabular}{@{}l rrrr rr r r@{}}
    \toprule
    & S$+$G$+$ & S$+$G$-$ & S$-$G$+$ & S$-$G$-$ & agreement & $\kappa$ & \multicolumn{2}{c}{either run, on flawed / $C_\mathrm{spec}$} \\
    \midrule
    Claude  & 58 & 9  & 12 & 82  & 0.87 & 0.73 & -- & -- \\
    GPT-5.4 & 65 & 5  & 10 & 81  & 0.91 & 0.81 & 72 & 8 \\
    Kimi    & 28 & 17 & 11 & 105 & 0.83 & 0.55 & 55 & 1 \\
    \bottomrule
  \end{tabular}
\end{table}

\paragraph{Reference bugs.}
The reference-bug verdict is stable across prompts for Claude and GPT-5.4, and moderately so for Kimi (Table~\ref{tab:consistency-runs}). Of the 81 flawed entries, Claude flags 73 in at least two of its three runs and 51 in all three.

\paragraph{Mismatches.}
Claude's \flagMISMATCH{} verdicts are stable as well: 12 of the 13 \flagMISMATCH{} entries receive that flag in at least two runs. The exception, entry~17, has a single \flagMISMATCH{} run, its other runs blaming Claude's own specifications. When both S and G report a mismatch, its direction agrees for Claude (10 of 10 entries) and GPT-5.4 (5 of 5), but only on 14 of 22 entries for Kimi.

\paragraph{Full outcomes.}
The complete outcome of a run is less stable than the bug verdict alone. Over five classes (accepted, \flagMISMATCH{}, reference bug, own-specification bug, failure), S and G agree with $\kappa = 0.64$, $0.68$, and $0.48$ for Claude, GPT-5.4, and Kimi; excluding entries with a failure in either run raises Kimi's to $0.60$.

\paragraph{Contradictions.}
Within the \speciso{} runs, only two entries receive, from the same stack, both a proved isomorphism and a reference-bug flag. On entry~95, GPT-5.4's simple specification reproduces the reference bug, which the test cases of the guided run then exposed. On entry~67, Claude's multiple specifications run proved a vacuous isomorphism (Section~\ref{app:consistency-provers}). A third case, Kimi on entry~160, pairs an isomorphism through the flawed helper with a reference-bug flag raised only at the proof stage.

\subsection{Across tasks}

\begin{table}[h]
  \caption{Reference-bug flags at the specification stage (S or G) and at the proof stage against the reference (PG).}
  \label{tab:consistency-tasks}
  \centering
  \small
  \begin{tabular}{@{}l rr r r r@{}}
    \toprule
    & \multicolumn{2}{c}{flagged entries (on flawed)} & S/G flags & PG flags on & \proofgen{} \flagOK{} \\
    \cmidrule(lr){2-3}
    & S or G & PG & also in PG & flawed entries & on flawed entries \\
    \midrule
    Claude  & 79 (--) & 44 (44) & 44/79 & 44/44 & 34 \\
    GPT-5.4 & 80 (72) & 41 (41) & 37/80 & 41/41 & 33 \\
    Kimi    & 56 (55) & 39 (39) & 32/56 & 39/39 & 24 \\
    \bottomrule
  \end{tabular}
\end{table}

\paragraph{Proof-stage flags.}
Reference bugs raised during \proofgen{} never contradict the annotation: for every stack, each one falls on a flawed entry, including the few that no \speciso{} run had flagged (4 for GPT-5.4, 7 for Kimi).

\paragraph{Coverage.}
The proof stage alone, however, surfaces only about half of the flawed references (Table~\ref{tab:consistency-tasks}). On most of the others, the stacks certify an implementation against the flawed specification: since the specification is visible at generation time, an implementation can satisfy it even when it misstates the intent. This happens more often when the bug weakens the specification. Taking as weakening the types \emph{missing constraint}, \emph{missing edge-case clause}, \emph{vacuous disjunct}, and \emph{trivial existential} (Appendix~\ref{app:taxonomy}), Claude certifies 13 of 17 such entries against 21 of the other 64 (one-sided Fisher $p = 0.0015$), GPT-5.4 13 against 20 ($p < 0.001$), and Kimi 8 against 16 ($p = 0.07$). This analysis is exploratory, as the direction is inferred from the bug type rather than annotated per entry. Specification-level comparison, rather than implementation proof, is thus what exposes most reference bugs.

\subsection{Across provers}
\label{app:consistency-provers}

\begin{table}[h]
  \caption{Reference-bug flags (S, G, or PG) against the merged annotation, over all 161 entries. Sensitivity over the 81 flawed entries; false positives (FP) over the 80 entries of $C_\mathrm{spec}$. Claude is omitted, since the annotation derives from its runs.}
  \label{tab:consistency-provers}
  \centering
  \small
  \begin{tabular}{@{}l r c r r r@{}}
    \toprule
    & sensitivity & 95\% CI & FP & $\kappa$ & Fisher $p$ \\
    \midrule
    GPT-5.4                  & 76/81 & [86.2, 98.0] & 8 & 0.84 & $4.7\cdot10^{-30}$ \\
    Kimi                     & 62/81 & [65.8, 85.2] & 1 & 0.75 & $3.0\cdot10^{-26}$ \\
    GPT-5.4 $\cup$ Kimi      & 77/81 & [87.8, 98.6] & 8 & 0.85 & $3.4\cdot10^{-31}$ \\
    \bottomrule
  \end{tabular}
\end{table}

\paragraph{Re-attribution.}
Counting flags from S, G, and PG, GPT-5.4 and Kimi flag 76 and 62 of the 81 flawed entries, and together 77 (95.1\%, Table~\ref{tab:consistency-provers}); from S and G alone, they flag 72 and 55 (Table~\ref{tab:consistency-runs}). 
All three stacks flag 60 of the flawed entries. Measured between stacks rather than against the annotation, agreement on the bug verdict is $\kappa = 0.81$ between Claude and GPT-5.4, $0.75$ between Claude and Kimi, and $0.72$ between GPT-5.4 and Kimi.

\paragraph{False positives.}
All 8 flags on $C_\mathrm{spec}$ come from GPT-5.4, one of them also raised by Kimi (entry~57). Five fall on \flagMISMATCH{} entries (0, 29, 30, 57, 87) and three on \flagOK{} entries (25, 96, 118), and none is corroborated by the flagging stack's own \proofgen{}, which certifies an implementation against the same reference. Most false positives thus sit at the boundary between \flagISSUE{} and \flagMISMATCH{}. Since neither Claude nor, with one exception, Kimi flags these entries, they point to GPT-5.4 being quicker than the other stacks to call a defensible reading a bug.

\paragraph{Accepted isomorphisms on flawed entries.}
No stack proves a flawed reference valid. Seven runs have an isomorphism accepted on a flawed entry, and inspection of each proof shows why: three generated specifications reproduce the reference bug (GPT-5.4 on entries~95 and~97, Kimi on entry~20); on entry~160, the specifications of Claude and Kimi reuse the flawed helper; and Claude's multiple specifications run proved two \emph{vacuous} isomorphisms, on entries~67 and~160, where both specifications are unsatisfiable on some input for every implementation. \clever{}'s isomorphism theorem, of the form $\forall f.\,(\forall x.\,P(f,x)) \leftrightarrow (\forall x.\,Q(f,x))$, accepts any such pair, so an accepted specification is not, on its own, evidence of validity.

\paragraph{Misses.}
On the 4 flawed entries flagged by neither GPT-5.4 nor Kimi (20, 97, 103, 104), every \speciso{} run returns \flagMISMATCH{}, blames the stack's own specification, fails, or reproduces the bug. Of Kimi's 19 misses, 18 receive at least one \flagMISMATCH{} reading instead. This boundary is also where the stacks diverge most: only 8 of the 13 \flagMISMATCH{} entries receive a \flagMISMATCH{} reading from all three, and GPT-5.4 flags 5 of them as reference bugs. Over the five outcome classes, pairwise $\kappa$ between stacks ranges from $0.39$ to $0.61$, against $0.72$--$0.81$ for the binary bug verdict.

\paragraph{Failures.}
Capability differences surface mainly as failures, and only rarely as contradictions: runs end with \flagFAIL{} or a timeout in 27, 59, and 170 of the 805 certification attempts (161 entries times S, G, PG, and the two proofs against generated specifications) for Claude, GPT-5.4, and Kimi.

\section{Taxonomy}
\label{app:taxonomy}
Autoformalization is hard to evaluate. Beyond producing a Lean specification that compiles, two substantive issues dominate: choosing the right \emph{strength} for the specification (loose enough not to over-constrain valid implementations, yet tight enough to rule out incorrect ones), and committing to a single \emph{interpretation} of an inherently ambiguous NL description. Problem~124, for instance, requires dates in the \texttt{mm-dd-yyyy} format but leaves implicit whether single-digit months and days should be accepted; and the ground-truth specification of problem~27 only requires the output characters to have the opposite case of the corresponding input ones, an uncommon but defensible reading of the prose.
In this sense, the central challenge in autoformalizing NL sources is that of \emph{formalizing the intent} behind them, bridging the gap between the informal human specification and its formal counterpart~\citep{lahiriIntentFormalizationGrand2026}: even direct human evaluation can leave subtle bugs unnoticed. Claude's downstream \proofgen{} attempts sometimes surfaced generated-specification defects that \speciso{} had missed, because a successful equivalence check alone does not guarantee fidelity to the NL task. The distinction between passing a formal check and formalizing the intended behavior is the main focus of the audit.

\begin{table}[h]
  \caption{Taxonomy of specification \flagISSUE{} types.}
  \label{tab:spec-issue-taxonomy}
  \centering
  \footnotesize
  \setlength{\tabcolsep}{4pt}
  \begin{tabularx}{\textwidth}{l X}
    \toprule
    \textbf{Type} & \textbf{Description} \\
    \midrule
    \emph{Vacuous disjunct}                  & $\lor$ (often a biconditional disjunct like \lean{result = none} $\leftrightarrow$ \lean{list = []}) trivially satisfied, admitting trivial implementations. \\
    \emph{Precedence bug}                    & Missing parentheses; $\land$/$\to$/$\leftrightarrow$ precedence makes the spec parse differently than intended (e.g.\ $\land$ binds tighter than $\to$, making the per-element constraint vacuous). \\
    \emph{Conjunct used as guard}            & \lean{(P $\land$ \dots)} used where \lean{(P $\to$ \dots)} was intended, making the spec unsatisfiable for invalid inputs (or, vice versa, vacuous). \\
    \emph{Trivial existential}               & $\exists$ or $\exists!$ trivially satisfied (e.g.\ \lean{$\exists$! S, S.card = 0} witnessed by $\emptyset$; \lean{$\exists$ v : $\mathbb{Q}$, |v - n| = 0} by \lean{v = $\uparrow$n}). \\
    \emph{Universal over invalid domain}     & Universal quantifier ranges over inputs where the body is unsatisfiable (e.g.\ over all \lean{Int n} requiring \lean{length = n} for negative \lean{n}; over all bases including~0). \\
    \emph{Missing edge-case clause}          & Missing clause for a particular edge case (\lean{n = 0}, empty list, single element, all-equal, all-odd, etc.). \\
    \emph{Missing constraint}                & Missing a non-edge constraint required by the NL description or test cases (positivity, range, ordering, format, multiplicity, sublist, stability for ties, etc.). \\
    \emph{Extra/over-restrictive constraint} & Extra restriction not in the NL description or test cases (e.g.\ extra \lean{result.data.all isLower} precondition, tolerance too tight). \\
    \emph{Nat arithmetic truncation}         & Lean \lean{Nat} truncation: subtraction floors at~0 and division floors, making the spec lossy. \\
    \emph{Out-of-bounds default access}      & \lean{getElem!}/\lean{get!} returns a default value out of bounds, weakening order, equality or index constraints. \\
    \emph{Type-system mismatch}              & \lean{Nat}/\lean{Int} coercion or other type-induced unsatisfiability. \\
    \emph{Wrong semantics}                   & Wrong formula, operation, predicate or filter, or structural mismatch (membership vs.\ count vs.\ sublist; midrange vs.\ median; backwards divisibility; off-by-one threshold; \lean{String.toLower} vs.\ \lean{Char.isLower}; etc.). \\
    \bottomrule
  \end{tabularx}
\end{table}

\paragraph{Ground-truth specifications.}
As stated in Section~\ref{sec:results}, 80 of the 161 ground-truth \probspec{} were flagged at least once as having bugs. We then had Claude classify the bugs arising in each \flagISSUE{} entry across the three \speciso{} runs and the \proofgen{} run from ground-truth specifications, obtaining a comprehensive taxonomy of 12 error types; for 5 of those problems (numbers 81, 99, 109, 130, 139), two types of issue were identified.
Most issues cluster on \emph{Lean-encoding} pitfalls: \emph{conjunct-as-guard} patterns using $(P \wedge Q)$ in place of $(P \to Q)$ (15 problems); \emph{precedence bugs} from omitted parentheses around $\wedge, \to, \leftrightarrow$ (12); universal quantifiers over inputs whose body is unsatisfiable (7); plus smaller counts of lossy \lean{Nat} arithmetic, default-value access via \lean{getElem!}/\lean{get!}, type-system mismatches, vacuous disjuncts, and trivial existentials. Together, these Lean-language hazards affect 48 of the 80 problems; the remaining 34 carry errors in the specification semantics: \emph{over-} or \emph{under-restrictive constraints} (10 problems), \emph{missing edge-case clauses} (4), and, most prominently, overall wrong interpretations of the problem departing from the NL description (21). See Table~\ref{tab:ground-truth-spec-errors}.

\begin{table*}[h]
  \caption{Frequencies by type of ground-truth specification issues identified across the Simple, Guided, Multiple runs of \speciso{} and the \proofgen{} run from ground-truth specs. Asterisks count the entries that were flagged as \flagISSUE{} in both specs (\speciso{}) or in spec and impl (\proofgen{}); superscripts mark entries classified under multiple types via the partner type's initial.}
  \label{tab:ground-truth-spec-errors}
  \centering
  \footnotesize
  \setlength{\tabcolsep}{3pt}
  \begin{tabularx}{\textwidth}{l c c c c X}
    \toprule
    \textbf{Type}                            & Simp.       & Guid.    & Multi          & \textsc{Proof}     & \textbf{Problem IDs} \\
    \midrule
    \emph{Conjunct used as guard}            & 14          & 13       & 29             & $9{+}1^{O}$        & 24, 44, 49, 50, 72, 81, 84, 110, 115, 122, 140, 150, 152, 158, 163 \\ %
    \emph{Precedence bug}                    & 10          & 11       & $29^{***}$     & 3                  & 23, 45, 51, 69, 75, 85, 88, 93, 109, 126, 132, 154 \\ %
    \emph{Universal over invalid domain}     & 7           & 6        & 16             & 5                  & 7, 67, 111, 125, 130, 143, 155 \\ %
    \emph{Nat arithmetic truncation}         & $3^{*}$     & 2        & 8              & 2                  & 36, 109, 139, 153 \\ %
    \emph{Out-of-bounds default access}      & 2           & 3        & 7              & $2{+}1^{C}$        & 26, 64, 66, 81 \\ %
    \emph{Type-system mismatch}              & 3           & 2        & 8              & 3                  & 100, 106, 151 \\ %
    \emph{Vacuous disjunct}                  & 3           & 3        & $7^{***}$      &                    & 12, 68, 161 \\ %
    \emph{Trivial existential}               & 1           & 2        & 5              &                    & 34, 39 \\ %
    \midrule
    \emph{Wrong semantics}                   & 18          & 19       & $38^{*}$       & $15^{**}$          & 27, 47, 65, 86, 89, 90, 91, 94, 95, 98, 99, 103, 108, 114, 116, 117, 121, 129, 130, 134, 144 \\ %
    \emph{Missing constraint}                & 1           & 5        & 12             & 2                  & 20, 79, 97, 99, 105, 138, 139, 159 \\ %
    \emph{Extra/over-restrictive constraint} & 2           & 1        & 5              & 1                  & 71, 156 \\ %
    \emph{Missing edge-case clause}          & 3           & 3        & 7              & 1                  & 5, 62, 78, 104 \\ %
    \bottomrule
  \end{tabularx}
\end{table*}

\paragraph{Generated specifications.}
In contrast, Claude-generated specifications fail mainly in a \emph{semantic} register, misstating the relation or under-specifying it, and rarely in a \emph{syntactic} one (only two instances: a \emph{trivial existential} and a \emph{type-system mismatch}). Within the semantic class, \emph{missing constraints} are more frequent than \emph{wrong semantics}: a different failure profile from the ground-truth specifications.
Per-run counts expose two phenomena. First, guidance is effective: the design guidelines instructed Claude on the policy for imposing constraints, producing a five-fold drop in issues (from 10 to 2) and the complete disappearance of \emph{missing-constraint} and \emph{over-/under-restrictive-constraint} errors. Second, those errors reappear in the multi-specification run despite the same guidelines: beyond the mechanical multiplication of surface area from more generated specifications per problem, this also reflects Claude's intentionally more ``adventurous'' generation, attempting to cover as much of the ``interpretation space'' as possible. See Table~\ref{tab:generated-spec-errors}.

\begin{table*}[h]
  \caption{Frequencies by type of Claude-generated specification \emph{semantic} issues only across \speciso{} runs as in Table~\ref{tab:ground-truth-spec-errors} and the \proofgen{} runs from generated specs (Simple and Guided prompts). Asterisks and single-letter superscripts formatted similarly to Table~\ref{tab:ground-truth-spec-errors}.}
  \label{tab:generated-spec-errors}
  \centering
  \footnotesize
  \setlength{\tabcolsep}{3pt}
  \begin{tabular}{@{}l c c c c c@{}}
    \toprule
    \textbf{Type}                            & Simple    & Guided  & Multiple                            & \textsc{Proof} ($\to$ Simp.) & \textsc{Proof} ($\to$ Guid.) \\
    \midrule
    \emph{Wrong semantics}                   & 2         & 2       & $10^{**}{+}3^{M\!*}$                & 5                             & 1                             \\ %
    \emph{Missing constraint}                & $5^{*}$   &         & $19^{****}{+}1^{E}{+}3^{W\!*}$      & 1                             &                               \\ %
    \emph{Extra/over-restrictive constraint} & 2         &         & $2{+}1^{M}$                         &                               &                               \\ %
    \bottomrule
  \end{tabular}
\end{table*}

\paragraph{Mismatches.}
The distribution of \flagMISMATCH{} entries, where neither specification is buggy but the two are not isomorphic, reflects \emph{authoring choices} rather than errors. Claude's generated specifications more often impose stronger constraints than the references, though guidance narrows this skew. Across the reviewed runs, 73 taxonomy assignments concern the boundary of the input domain (gaps in preconditions, edge-case handling); 37 concern conditions imposed on the function output (ordering of lists vs.\ multisets, canonical-form impositions); 22 concern structural and definitional choices; and a minority of 10 are semantic in nature, e.g.\ different but mathematically defensible formulations, or quantification over indices instead of values. An entry can receive multiple assignments. See Table~\ref{tab:mismatch-frequency}.

\begin{table*}[h]
  \caption{Mismatch frequencies by type and direction across the \speciso{} runs. Direction symbols: $>$ generated stronger than reference, $/$ neither, $<$ generated weaker. Single-letter superscripts are formatted similarly to Table~\ref{tab:ground-truth-spec-errors}.}
  \label{tab:mismatch-frequency}
  \centering
  \footnotesize
  \setlength{\tabcolsep}{4pt}
  \begin{tabular}{l c c c c c c c c c}
    \toprule
    & \multicolumn{3}{c}{Simple} & \multicolumn{3}{c}{Guided} & \multicolumn{3}{c}{Multiple} \\
    \cmidrule(lr){2-4} \cmidrule(lr){5-7} \cmidrule(lr){8-10}
    \textbf{Type}                    & $>$ & $/$               & $<$ & $>$ & $/$                       & $<$ & $>$                  & $/$                         & $<$         \\
    \midrule
    \emph{Precondition gap}          & 2  & $1^{O}$           & 1   & 7   & $2^{S}{+}1^{T}$           & 5   & $11{+}1^{S}{+}1^{T}$ & $2{+}3^{O}{+}2^{C}{+}2^{S}$ & $7{+}1^{S}$ \\ %
    \emph{Edge-case behavior}        & 4  &                   &     & 5   &                           & 2   & $2{+}1^{A}$          & 1                           & 9           \\ %
    \emph{Ordering vs multiset}      & 5  & $1^{P}$           &     & 4   &                           &     & 10                   & $3^{P}{+}1^{S}$             & 2           \\ %
    \emph{Canonical form}            & 1  &                   &     & 1   & 1                         &     & 3                    & $3{+}2^{P}$                 &             \\ %
    \emph{Structural-op choice}      & 1  & 1                 &     &     & $1{+}2^{P}$               &     & $1{+}1^{P}$          & $6{+}2^{P}{+}1^{O}$         & $1^{P}$     \\ %
    \emph{Type-system semantics}     &    & $2{+}1^{A}$       &     &     & $1^{P}$                   &     & $1^{P}$              &                             &             \\ %
    \emph{Algorithmic variant}       & 1  & $1^{T}$           &     &     & 1                         &     & $1{+}1^{E}$          &                             &             \\ %
    \emph{Quantifier scope}          & 1  &                   &     &     &                           & 1   & 2                    &                             &             \\ %
    \emph{Adjacency}                 &    & 1                 &     &     &                           &     &                      &                             &             \\ %
    \bottomrule
  \end{tabular}
\end{table*}

\begin{table}[h]
  \caption{Taxonomy of specification \flagMISMATCH{} types.}
  \label{tab:mismatch-taxonomy}
  \centering
  \footnotesize
  \setlength{\tabcolsep}{4pt}
  \begin{tabularx}{\textwidth}{l X}
    \toprule
    \textbf{Type} & \textbf{Description} \\
    \midrule
    \emph{Edge-case behavior}        & The disagreement is on a small, finite or enumerable set of edge inputs (\lean{n=0}; empty list; singleton; \lean{p=0} in division; min~=~max; all-equal). The natural fix is to add 1--2 specific case clauses. \\
    \emph{Precondition gap}          & The disagreement is on a broad or infinite class of inputs characterised by a property (\lean{n < 0}; \lean{length > 1000}; non-paren chars; non-digit chars; invalid operator; non-lowercase output). The natural fix is to remove or generalise a domain restriction. \\
    \emph{Quantifier scope}          & Different quantifier domain: $\forall$ over distinct values vs.\ distinct indices; existential parsing vs.\ universal verification; $\exists!$ over a set vs.\ a sum over a multiset. \\
    \emph{Ordering vs multiset}      & One spec requires a specific order, exact list, or \lean{result = list.filter \dots}; the other allows permutation, multiset, set, or membership semantics. \\
    \emph{Canonical form}            & One spec requires a unique canonical representation; the other allows equivalents (no leading zeros, normalised form, fixed-width vs.\ variable-width formatting, single vs.\ multiple representations of the same value). \\
    \emph{Structural-op choice}      & Different filter, split or recursion choices (\lean{split} vs.\ \lean{splitOn}~+~filter; with vs.\ without empty-token filtering; preserving vs.\ collapsing duplicates) that diverge on edge inputs. \\
    \emph{Type-system semantics}     & Different choice of Lean type-system semantics: \lean{Int.mod} Euclidean vs.\ \lean{Nat.mod} vs.\ \lean{natAbs}; \lean{Rat}-field vs.\ \lean{Int}-integer arithmetic; \lean{Nat}-truncation vs.\ \lean{Int}-exact; \lean{String.toLower} vs.\ \lean{Char.isLower}. \\
    \emph{Algorithmic variant}       & Different but mathematically defensible formulation: binary digit sum vs.\ decimal digit sum; midrange vs.\ median; threshold \lean{n $\geq$ 2} vs.\ \lean{n > 2}; different mathematical algorithm. \\
    \emph{Adjacency}                 & Adjacent or consecutive positions vs.\ arbitrary indices (consonant--vowel--consonant requiring \lean{(i-1, i, i+1)} vs.\ \lean{i < j < k}). \\
    \bottomrule
  \end{tabularx}
\end{table}

\paragraph{Implementations.}
Claude detected 7 issues in the generated implementations, on only 5 problems, across the three \proofgen{} runs. Three involve hard-coded \emph{bounded fuel} (e.g.\ the 200-iteration Newton--Raphson loop in problem~32, see Appendix~\ref{app:notable-examples}) that does not suffice for the arbitrarily large inputs that the specification quantifies over; the remaining four are off-by-one indexing errors (problems~47 and~163), a missing case (a stationary point for Newton's method in problem~32 again), and a single wrong-algorithm choice (problem~116: the implementation sorts numerically, while the specification requires sorting by binary digit count).

\begin{table}[h]
  \caption{Taxonomy of implementation \flagISSUE{} types.}
  \label{tab:impl-issue-taxonomy}
  \centering
  \footnotesize
  \setlength{\tabcolsep}{4pt}
  \begin{tabularx}{\textwidth}{l X}
    \toprule
    \textbf{Type} & \textbf{Description} \\
    \midrule
    \emph{Bounded fuel}                  & Implementation uses a fixed iteration count or fuel insufficient for arbitrary inputs (e.g.\ 200 Newton iterations, 64 power checks, 1000 Collatz steps). Fixable by adding more fuel or unbounded recursion. \\
    \emph{Off-by-one}                    & Index or range off-by-one error. \\
    \emph{Nat truncation}                & \lean{Nat} subtraction floors at~0 or division truncates, causing wrong implementation behavior. \\
    \emph{Out-of-bounds default access}  & Implementation uses \lean{get!}/\lean{getElem!} returning a default value outside bounds. \\
    \emph{Wrong algorithm}               & Implementation encodes a fundamentally different algorithm than the spec demands. \\
    \emph{Missing case}                  & Implementation does not handle some edge case (e.g.\ stationary point in Newton's method). \\
    \emph{Wrong formula}                 & Wrong arithmetic or formula in the implementation. \\
    \bottomrule
  \end{tabularx}
\end{table}

\paragraph{Benchmark.}
Four problems carry issues in the fixed, non-editable parts of the benchmark itself, which neither the generated specification nor the generated implementation can repair. Problem~123 asks for a \texttt{collatz} iterator: any proof of correctness against the formal specification essentially requires a proof of the Collatz conjecture itself; problem~39 (\texttt{prime\_fib}) similarly hinges on the open question of the existence of infinitely many Fibonacci primes; and problem~160 (\texttt{do\_algebra}) ships a fixed helper \lean{applyOp} that returns a \lean{none} on division by zero, which is not parsed by other parts of the script.
A notable example, Problem 32, exhibits a missing rational-root precondition, and highlights other issues and phenomena observed in the results; it is discussed in detail in Appendix~\ref{app:notable-examples}.

\section{Resources}
\label{app:resources}
\label{app:runtimes-and-costs}

This appendix reports how the three stacks spent time, tokens, and money. Formalization tasks are \specgen{} and \implgen{}; proving tasks are \speciso{} and \proofgen{}.

\paragraph{Time.}
Wall-clock time was spent mostly on model inference, not on client-side work such as tool calls, Lean builds, and LSP queries (Table~\ref{tab:resources-time}). On proving tasks, tool execution took 20\% of the total wall-clock time for Claude and GPT-5.4, and 6\% for Kimi; on formalization tasks, 12\%, 29\%, and 22\%. For GPT-5.4 and Kimi, whose traces time each tool call, nearly all of it is Lean tooling. Inference speed thus set the pace, and it differed with the serving: Claude and GPT-5.4 ran through their vendors' APIs, Kimi on a local cluster. On proving tasks, a median Kimi turn took 27\,s against 12\,s for Claude and 7\,s for GPT-5.4, and Kimi also needed more turns. Its median proving attempt lasted 47 minutes, against 7 and 6 minutes for the other two, which is why it received a 4-hour budget instead of 1 hour; even so, 9.6\% of its proving attempts reached the timeout. Formalization tasks took about a minute and a half on every stack.

\begin{table}[h]
  \caption{Time per attempt, by stack and task family: median wall-clock time, share of wall-clock time spent executing tools (summed over attempts), and, for proving tasks, median turns, median inference seconds per turn, and share of attempts reaching the timeout (1\,h for Claude and GPT-5.4, 4\,h for Kimi).}
  \label{tab:resources-time}
  \centering
  \small
  \begin{tabular}{@{}l rr rr rrr@{}}
    \toprule
    & \multicolumn{2}{c}{median wall-clock} & \multicolumn{2}{c}{tool share} & \multicolumn{3}{c}{proving} \\
    \cmidrule(lr){2-3} \cmidrule(lr){4-5} \cmidrule(l){6-8}
    & formal. & proving & formal. & proving & turns & s/turn & timeouts \\
    \midrule
    Claude  & 83\,s & 6.7\,min & 12\% & 20\% & 27 & 11.7 & 2.8\% \\
    GPT-5.4 & 85\,s & 5.6\,min & 29\% & 20\% & 36 & 6.9  & 0\% \\
    Kimi    & 91\,s & 47\,min  & 22\% & 6\%  & 83 & 26.6 & 9.6\% \\
    \bottomrule
  \end{tabular}
\end{table}

\begin{figure}[h!]
  \centering
  \includegraphics[width=\textwidth]{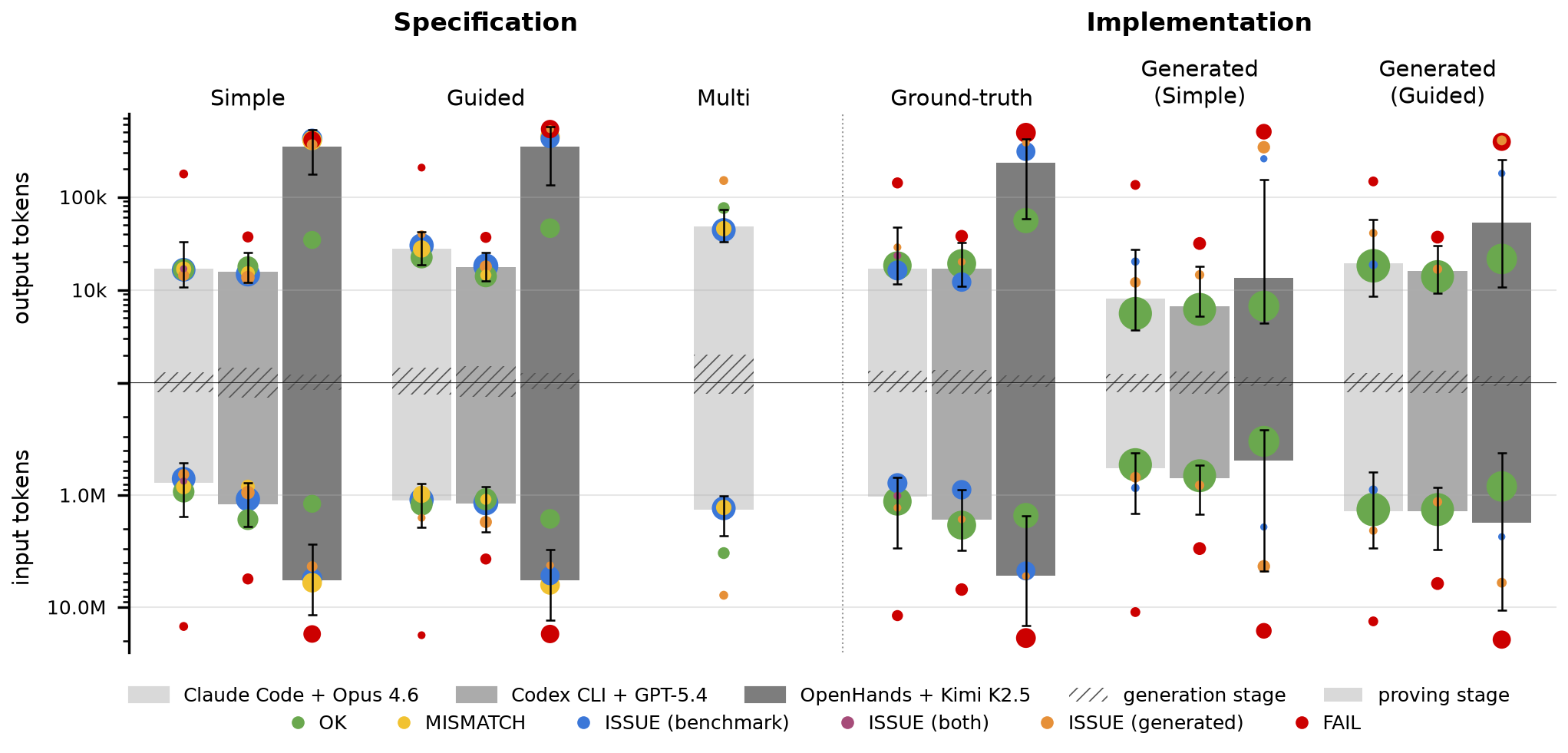}
  \caption{Median tokens per attempt, by stack and prompt regime, for specification certification (\specgen{} and \speciso{}) and implementation certification (\implgen{} and \proofgen{}, against the reference or against the stack's simple or guided generated specification). Output tokens grow upward and input tokens, including cached ones, downward, each on its own logarithmic scale; error bars span the 25th to 75th percentiles. The hatched segment near the axis shows the generation stage's share of the total, not its absolute count. Dots mark the median of each outcome class, sized by its frequency. The multiple specifications run (Claude only) covers 105 entries. Kimi's endpoint does not report cached tokens; nine Claude records without complete token counts are excluded.}
  \label{fig:token-bars}
\end{figure}

\paragraph{Tokens.}
Token usage mirrors time (Figure~\ref{fig:token-bars}). A typical attempt of Claude or GPT-5.4 reads between 0.6 and 1.7 million input tokens, mostly from the prompt cache, and writes 7 to 28 thousand output tokens (medians over the shared prompt regimes), with the generation stage accounting for a small share of both. Kimi's medians are several times higher on the specification runs and on proofs against the reference (5.8 and 5.3 million input, 350 and 235 thousand output tokens), but the gap comes from its failures, not its successes. Its successful attempts use about as many tokens as those of the other stacks (0.3 to 1.9 million in total, depending on the task), while its failed attempts, which run until the 4-hour budget and make up 15--27\% of its runs, consume 17 to 20 million each. Failed attempts are the most token-intensive outcome for every stack.

\begin{figure}[h!]
  \centering
  \includegraphics[width=\textwidth, height=\textheight, keepaspectratio]{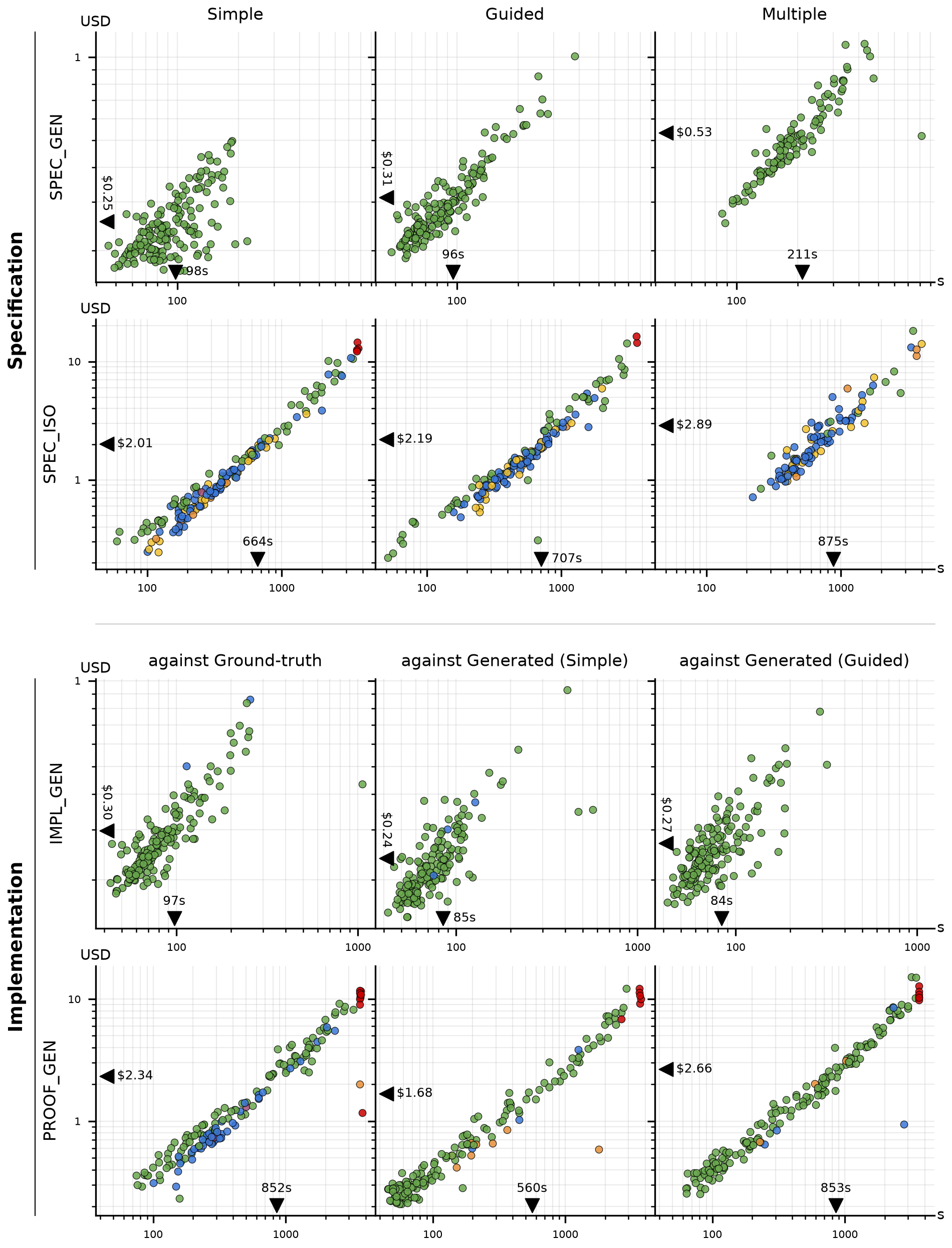}
  \caption{Cost (USD) against runtime (s) of Claude's attempts, across all tasks and runs. Markers follow the color scheme of Figure~\ref{fig:claude-flags}; labeled black triangles mark the average cost and runtime of each run.}
  \label{fig:resources-costs}
\end{figure}

\paragraph{Costs.}
Only the Claude runs report costs: the GPT-5.4 harness did not provide them, and the local cluster serving Kimi is not billed per token. As reported by the Claude Agent SDK, costs per attempt grew with runtime (Figure~\ref{fig:resources-costs}) and, more tightly, with token usage: a median of \$0.25 for formalization tasks (\$0.49 for the multiple specifications run) and \$1.17 for proving tasks, whose mean of \$2.26 reflects a long tail of slow attempts. Attempts that reached the 1-hour timeout cost about \$11 each.

\clearpage
\section{Contamination}
\label{app:contamination}
\clever{} was public well before the release of the models in our three stacks: its repository since February 2025 and its paper since May 2025. Training exposure to the benchmark therefore cannot be fully excluded, and it is impossible to measure its effect on any stack directly. We argue instead that the audit's findings do not depend on it, on three grounds, and report an exploratory control with a model trained before \clever{}'s release.

\paragraph{Little formal material to memorize.}
The algorithmic content of \clever{} is public by construction: the \humaneval{} problems~\citep{chen2021evaluatinglargelanguagemodels} and their Python solutions have circulated since 2021, and the benchmark ships reference solutions in Python and Rust. Its formal layer offers far less. Besides the 161 reference specifications, it includes Lean reference implementations for 43 entries and complete correctness proofs for only 4 (entries 0, 3, 17, and 72); no isomorphism proof exists for any entry. Most \speciso{} and \proofgen{} targets thus have no public solution to recall.

\paragraph{Memorization would work against the audit.}
A model that recalled the public reference specifications would reproduce them: its specifications would be trivially isomorphic to the references and would inherit their bugs. We observe the opposite. Only 65 of 161 entries end with an \emph{accepted} specification in the merged annotation, and 81 references are flagged as flawed, each with a supporting argument, typically a concrete counterexample, that we can check independently of how the model found it. Exposure could make \emph{accepted} specifications easier to obtain; it cannot explain why the models reject the references they would have memorized.

\paragraph{Independent replication.}
GPT-5.4 and Kimi, which share neither model nor agent framework with Claude, re-attribute to the reference 77 of the 81 flaws (Appendix~\ref{app:consistency}). A flaw found by several independently trained models, and certified by a machine-checkable argument, is a property of the benchmark rather than of one model's training data.

\paragraph{Pre-cutoff control.}
A direct control requires a model trained before \clever{} became public that is also strong enough in our agentic setup for its score to be informative: a weak model scoring low cannot distinguish the absence of contamination from the lack of capability. No currently served Claude model meets the first requirement. We first ran Qwen3-235B-A22B, with training cutoff likely predating the release of \clever{}'s repository, in the OpenHands harness, but the model proved too weak to be informative and we abandoned it.

\begin{figure}[h]
  \centering
  \includegraphics[scale=0.73]{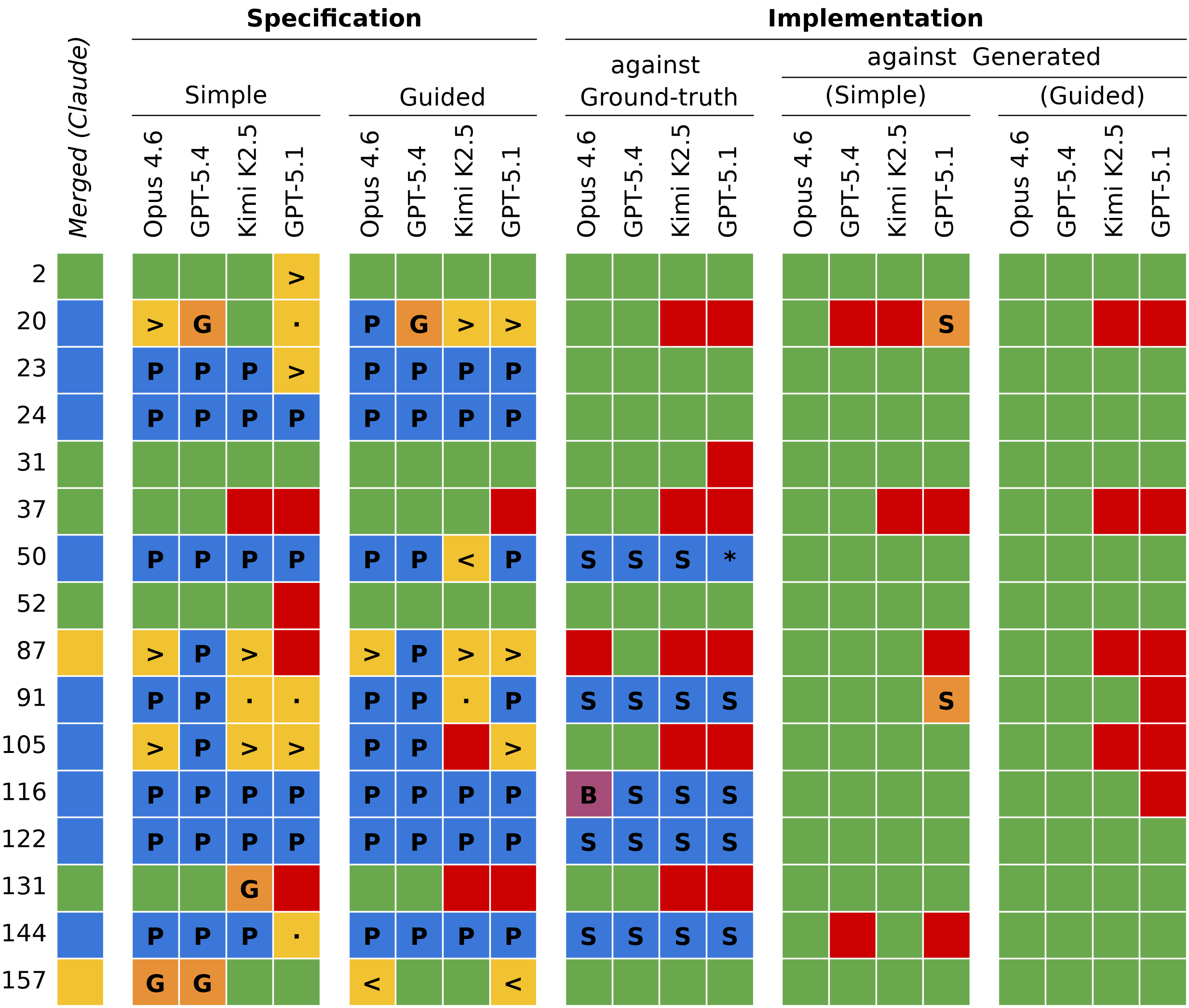}
  \caption{Per-entry outcomes of the 16-entry pilot for the three stacks and the pre-cutoff GPT-5.1, over the simple and guided \speciso{} runs, \proofgen{} against the reference, and \proofgen{} against the simple and guided generated specifications (last two groups). The leftmost column is Claude's merged annotation. Colors and symbols follow Figure~\ref{fig:claude-flags}.}
  \label{fig:models-grid-pre}
\end{figure}

We then turned to GPT-5.1, whose vendor-stated knowledge cutoff (September 30, 2024) predates \clever{}'s release, while it belongs to the same model family as GPT-5.4. We ran it on all ten tasks of a 16-entry pilot (10\% of dataset), drawn uniformly at random with a fixed seed, in the same Codex CLI harness, with the same prompts and settings as GPT-5.4. The pilot contains 9 flawed entries, 5 annotated \flagOK{}, and 2 \flagMISMATCH{} (Figure~\ref{fig:models-grid-pre}).
\begin{itemize}[leftmargin=*,itemsep=1pt,topsep=2pt]
  \item \textbf{GPT-5.1 reproduces the audit's central finding.} It flags the reference as flawed on 7 of the 9 flawed entries and on none of the other 7 (Cohen's $\kappa = 0.75$ against the annotation; one-sided Fisher $p = 0.003$). Its two misses (entries 20 and 105) are \flagMISMATCH{} readings, and it never proves a flawed reference valid.
  \item \textbf{GPT-5.1 closes fewer proofs.} It succeeds on 31 of the 80 certification attempts (16 entries $\times$ five certification runs) against 52 for GPT-5.4, which is ahead on 10 entries and behind on none (sign test $p = 0.002$); end-to-end success is 11 against 15 of 16. %
\end{itemize}

None of these arguments proves the absence of contamination, and exposure could still help the main stacks, for instance by making the references' formulation style familiar. Such help would raise the counts of \emph{accepted} specifications and certified implementations; it would not produce the flaws, mismatches, and counterexamples on which the audit rests.

\clearpage
\section{A Notable Example: Problem~32}
\label{app:notable-examples}
Problem~32 deserves a separate look, both because it is arguably the hardest problem in the dataset and because it gathers in a single entry several of the phenomena observed throughout our evaluation. The task asks to find a real root of an odd-degree polynomial given by its list of coefficients.

\begin{figure}[h]
 \centering
 \includegraphics[width=0.95\textwidth]{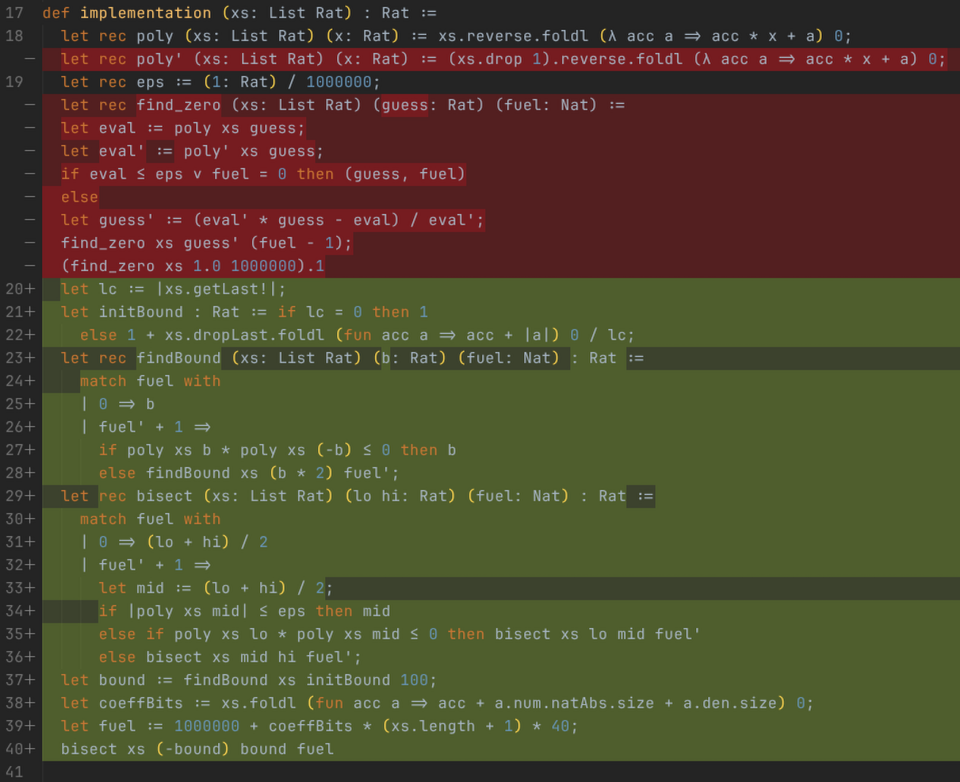}
 \caption{Claude Code's fix of the reference implementation of problem~32 in the preliminary experiments: bisection from the Cauchy root bound replaces Newton--Raphson.}
 \label{fig:claude_fix}
\end{figure}

Already in the preliminary experiments (Appendix~\ref{app:preliminary-experiments}), the provided ground-truth implementation was found to be incorrect: it was based on the Newton--Raphson method, which is not globally convergent and can enter cycles even on simple polynomials. Following insight from the agent, it was replaced with a bisection method initialized from the Cauchy root bound, with bisection fuel adaptive in the bit-size of the coefficients, ensuring convergence to the target precision (see Figures~\ref{fig:claude_fix} and~\ref{fig:claude_extract}).

\begin{figure}[h]
 \begin{terminalbox}
  Here's a summary of the fix and why it's correct:\\

  \textcolor{termblue}{Root cause:} Newton-Raphson is not globally convergent. Even with a correct derivative, it can enter cycles (as shown by \textcolor{termpurp}{[2, -2, 0, 1]} cycling between 0 and 1). No starting point is safe for all polynomials.\\

  \textcolor{termblue}{Fix:} Replaced Newton-Raphson with the \textcolor{termblue}{bisection method}, which is globally convergent:\\
  1. \textcolor{termpurp}{findBound}: Finds a bound \textcolor{termpurp}{b} (starting from 1, doubling up to 2$^\text{100}$) such that \textcolor{termpurp}{poly(b)} and \textcolor{termpurp}{poly(-b)} have opposite signs. This is guaranteed to exist for odd-degree polynomials (even-length coefficient list) with non-zero leading
  coefficient -- exactly the docstring's preconditions.\\
  2. \textcolor{termpurp}{bisect}: Standard bisection on \textcolor{termpurp}{[-bound, bound]} with 1,000,000 iterations. Each step halves the interval containing the root. After n steps, the interval has width \textcolor{termpurp}{2·bound / 2$^\text{n}$}, giving astronomically precise convergence -- far
  more than needed for \textcolor{termpurp}{eps = 10$^\text{-6}$}.\\

  \textcolor{termblue}{Verification against test cases}:\\
  - \textcolor{termpurp}{[1, 2]} (1 + 2x): bisect finds mid = -0.5 where poly = 0 exactly in 2 steps\\
  - \textcolor{termpurp}{[-6, 11, -6, 1]} (x³-6x²+11x-6): bisect converges to a value near root 1 with |poly| $\leq$ eps

 \end{terminalbox}
 
 \begin{terminalbox}
  The fix has two changes to the implementation (spec unchanged):\\

  \textcolor{termblue}{Fix 1 - Cauchy bound initialization (lines 20-22)}: Instead of \textcolor{termpurp}{findBound xs 1 100}, we compute the Cauchy root bound \textcolor{termpurp}{1 + $\sum$|a$_\text{i}$|/|a$_\text{i}$|} as the starting value. This guarantees the initial bound is already past all roots, regardless of coefficient magnitude.\\

  \textcolor{termblue}{Fix 2 - Adaptive bisection fuel (lines 38-39)}: The bisection fuel scales with the total bit-size of the coefficients (\textcolor{termpurp}{coeffBits * (xs.length + 1) * 40 + 1000000}). This ensures enough precision even for polynomials with large/steep coefficients.

 \end{terminalbox}

 \caption{Extract of Claude Code's replies explaining its fix of problem~32.}
 \label{fig:claude_extract}
\end{figure}

At specification certification, no run produced a \genspec{} that proved isomorphic to the ground-truth one. The reason was systematic: every generated specification required the implementation to return an \emph{exact} root of the polynomial, whereas the ground-truth specification only required an \emph{approximate} root within a fixed tolerance. The \specgen{} agent could not have inferred this from context, since neither the NL description nor the example usages mentioned approximation or any threshold for it. The corresponding entries were flagged either as \flagMISMATCH{} or as \flagISSUE{} in \genspec{}.

At implementation certification, no run succeeded either. Two factors compound: the intrinsic difficulty of the implementation (global convergence of the root-finding method, Newton--Raphson vs.\ bisection, finite fuel), already witnessed by the fix to the ground-truth implementation; and the function signatures imposed by the benchmark, which require a rational return type to side-step the fact that \lean{Real} is not a computable type in Lean (as a developer comment in the entry itself acknowledges). Odd-degree polynomials such as $x^3 + 2$ have a real root but no rational one, so under the imposed \lean{Rat} type the task has no exact solution: any implementation can at best approximate the root. The \proofgen{} attempts were accordingly flagged as \flagISSUE{} in either the implementation or the benchmark-imposed signatures. The same observation explains why logically correct specifications were flagged as \flagISSUE{} at \speciso{} time: the agent recognized that any specification asking for an exact root must fail on polynomials without rational roots, regardless of the implementation~--- a trap from which the ground-truth specification escapes precisely by asking for an approximate root.

Problem~32 thus condenses several of the difficulties this kind of benchmark exposes: a literal reading of the NL description leads to a specification incompatible with the structure imposed by the benchmark; reconciling the two requires a creative workaround (here, an approximation threshold); and the implementation itself calls for non-trivial mathematical insight.

\clearpage
\section{Complete Finding Summaries}
\label{app:complete-findings}
\begingroup
\footnotesize
\setlength{\tabcolsep}{4pt}
\renewcommand{\arraystretch}{1.06}
\setlength{\LTpre}{8pt}
\setlength{\LTpost}{8pt}
This appendix lists the entry-level findings behind the aggregate counts in
Section~\ref{sec:discussion}. Each row gives the benchmark entry identifier,
the taxonomy term used in this paper, and a brief summary.
The full description of each finding, with its supporting argument or counterexample
as submitted for verification, is part of our release (\repourl{}).
Entries 32 and 39 have distinct findings in two categories; their
summaries are split accordingly. Entry 160 concerns a fixed helper
rather than the ground-truth specification, so it appears only in the last table.

\begin{longtable}{@{}>{\raggedright\arraybackslash}p{0.30in}>{\raggedright\arraybackslash}p{1.45in}>{\raggedright\arraybackslash}p{3.50in}@{}}
\caption{Ground-truth specification \flagISSUE{} findings (80 entries).}\label{tab:all-spec-issues}\\
\toprule
\textbf{ID} & \textbf{Taxonomy} & \textbf{Summary} \\
\midrule
\endfirsthead
\caption[]{Ground-truth specification \flagISSUE{} findings (80 entries). (continued)}\\
\toprule
\textbf{ID} & \textbf{Taxonomy} & \textbf{Summary} \\
\midrule
\endhead
\midrule
\multicolumn{3}{r}{\emph{Continued on next page}}\\
\endfoot
\bottomrule
\endlastfoot
5 & Missing edge-case clause & Singleton branch wrongly appends delimiter to lone element \\
\addlinespace[2pt]
7 & Universal over invalid domain & Guarded universal vacuously true, empty output satisfies spec \\
\addlinespace[2pt]
12 & Vacuous disjunct & \ensuremath{\lor} instead of \ensuremath{\land} voids the longest-element constraint \\
\addlinespace[2pt]
20 & Missing constraint & Minimality over distinct values misses equal-number zero-gap pairs \\
\addlinespace[2pt]
23 & Precedence bug & \ensuremath{\land} binds tighter than \ensuremath{\leftrightarrow}, length recurrence never enforced \\
\addlinespace[2pt]
24 & Conjunct used as guard & Divisor requirement sits as guard, vacuous when result=0 \\
\addlinespace[2pt]
26 & Out-of-bounds default access & Unbounded order-witness indices let getElem! default admit fabricated elements \\
\addlinespace[2pt]
27 & Wrong semantics & Constrains only output case category, not character identity \\
\addlinespace[2pt]
34 & Trivial existential & count=1 inside biconditional lets empty list satisfy \\
\addlinespace[2pt]
36 & Nat arithmetic truncation & Nat subtraction saturates, so 7-count tally can decrease \\
\addlinespace[2pt]
39 & Trivial existential & n=1 vacuous finset imposes no minimality \\
\addlinespace[2pt]
44 & Conjunct used as guard & Base-validity guard a conjunct not precondition, fails at base 0 \\
\addlinespace[2pt]
45 & Precedence bug & Missing parens make area constraint vacuous when a \ensuremath{\neq} 0 \\
\addlinespace[2pt]
47 & Wrong semantics & Even case computes midrange not median via swapped min/max sublists \\
\addlinespace[2pt]
49 & Conjunct used as guard & Modulus positivity a conjunct not guard, unsatisfiable at p=0 \\
\addlinespace[2pt]
50 & Conjunct used as guard & isAlphabetic s a conjunct not guard, fails on non-alphabetic input \\
\addlinespace[2pt]
51 & Precedence bug & Precedence guards consonant-preservation behind result.contains, empty string passes \\
\addlinespace[2pt]
62 & Missing edge-case clause & Length xs.length-1 clause contradicts length-preserving derivative equality \\
\addlinespace[2pt]
64 & Out-of-bounds default access & Empty string [0]! defaults to vowel A, making count unsatisfiable \\
\addlinespace[2pt]
65 & Wrong semantics & Reverse branch tests input palindrome instead of forcing reversed result \\
\addlinespace[2pt]
66 & Out-of-bounds default access & Empty string hits else, getElem! default forces 0=65 \\
\addlinespace[2pt]
67 & Universal over invalid domain & No precondition guards malformed input, existential unsatisfiable \\
\addlinespace[2pt]
68 & Vacuous disjunct & Never asserts result length in \{0,2\}, admits wrong list \\
\addlinespace[2pt]
69 & Precedence bug & -1 sentinel fused into biconditional, constant -1 satisfies spec \\
\addlinespace[2pt]
71 & Extra/over-restrictive constraint & 1/10000 tolerance on result\textsuperscript{2} too tight for 2-decimal rounding \\
\addlinespace[2pt]
72 & Conjunct used as guard & Only implications no converse, so false always satisfies \\
\addlinespace[2pt]
75 & Precedence bug & \ensuremath{\to} binds tighter than \ensuremath{\leftrightarrow}, misparses guard, fails at 101 \\
\addlinespace[2pt]
78 & Missing edge-case clause & 0<length guard leaves empty string unconstrained, misses base case \\
\addlinespace[2pt]
79 & Missing constraint & Never constrains inner digits to be binary bits \\
\addlinespace[2pt]
81 & Conjunct used as guard\newline Out-of-bounds default access & Range check as conjunct, result indexed by dict position \\
\addlinespace[2pt]
84 & Conjunct used as guard & Binary-character check placed as guard not conjunct, spec vacuous \\
\addlinespace[2pt]
85 & Precedence bug & Missing parens chain contradictory length guards, spec vacuously true \\
\addlinespace[2pt]
86 & Wrong semantics & Char count indexes each word separately, permutation never enforced \\
\addlinespace[2pt]
88 & Precedence bug & Missing parens demand sum odd and even at once, vacuous \\
\addlinespace[2pt]
89 & Wrong semantics & Shift direction inverted, spec decrypts instead of encrypting \\
\addlinespace[2pt]
90 & Wrong semantics & Second-smallest result never required to be list member \\
\addlinespace[2pt]
91 & Wrong semantics & Recursion double-counts final sentence and drops first \\
\addlinespace[2pt]
93 & Precedence bug & Missing parens make case-swap arms vacuous, identity passes \\
\addlinespace[2pt]
94 & Wrong semantics & Inequality picks smallest prime instead of largest \\
\addlinespace[2pt]
95 & Wrong semantics & Digit keys pass both toLower and toUpper checks \\
\addlinespace[2pt]
97 & Missing constraint & Missing natAbs makes negative operands' unit digit wrong \\
\addlinespace[2pt]
98 & Wrong semantics & Zero-count clause quantifies all indices not just even \\
\addlinespace[2pt]
99 & Wrong semantics\newline Missing constraint & Truncates instead of rounding; digitless string counts valid \\
\addlinespace[2pt]
100 & Type-system mismatch & result.length coerced to Int unsatisfiable for negative n \\
\addlinespace[2pt]
103 & Wrong semantics & Nat floor division instead of round-to-nearest average \\
\addlinespace[2pt]
104 & Missing edge-case clause & Zero has no digits so wrongly retained \\
\addlinespace[2pt]
105 & Missing constraint & Missing count constraint lets deduplicating implementation pass \\
\addlinespace[2pt]
106 & Type-system mismatch & result.length coerced to Int unsatisfiable for negative n \\
\addlinespace[2pt]
108 & Wrong semantics & Negates least-significant digit not most-significant \\
\addlinespace[2pt]
109 & Precedence bug\newline Nat arithmetic truncation & \ensuremath{\land}/\ensuremath{\leftrightarrow} precedence misgroups empty guard; Nat subtraction breaks rotation \\
\addlinespace[2pt]
110 & Conjunct used as guard & One-way guard lets constant YES satisfy spec \\
\addlinespace[2pt]
111 & Universal over invalid domain & Membership clause nested under keys, empty result vacuously satisfies \\
\addlinespace[2pt]
114 & Wrong semantics & List.sublists enumerates subsequences not contiguous subarrays \\
\addlinespace[2pt]
115 & Conjunct used as guard & Biconditional forbids zero result on non-empty all-zero grids \\
\addlinespace[2pt]
116 & Wrong semantics & Sorts by lexicographic binary digits instead of popcount \\
\addlinespace[2pt]
117 & Wrong semantics & splitOn empty words unfiltered makes spec unsatisfiable \\
\addlinespace[2pt]
121 & Wrong semantics & Reads lst[i+1]! summing odd positions instead of even \\
\addlinespace[2pt]
122 & Conjunct used as guard & \ensuremath{\land} instead of \ensuremath{\lor} forces qualifying element, unsatisfiable \\
\addlinespace[2pt]
125 & Universal over invalid domain & Minimality quantified over wrong-length paths makes spec unsatisfiable \\
\addlinespace[2pt]
126 & Precedence bug & \ensuremath{\land} binds tighter than \ensuremath{\to} collapsing chain to True \\
\addlinespace[2pt]
129 & Wrong semantics & words-none biconditional inverted and ord uses first-char codepoint \\
\addlinespace[2pt]
130 & Universal over invalid domain\newline Wrong semantics & Length n instead of n+1 contradicts at n=0 \\
\addlinespace[2pt]
132 & Precedence bug & \ensuremath{\to} binds tighter than \ensuremath{\leftrightarrow} mis-grouping the biconditional \\
\addlinespace[2pt]
134 & Wrong semantics & String.join drops spaces collapsing multi-word tail to one word \\
\addlinespace[2pt]
138 & Missing constraint & Missing positivity lets even zeros satisfy the sum \\
\addlinespace[2pt]
139 & Nat arithmetic truncation & Nat division makes factorial recurrence lossy and non-unique \\
\addlinespace[2pt]
140 & Conjunct used as guard & Non-maximal pref lets implementation dodge the dash collapse \\
\addlinespace[2pt]
143 & Universal over invalid domain & Empty result forces empty string into words, unsatisfiable \\
\addlinespace[2pt]
144 & Wrong semantics & Divisibility direction inverted between numerator and denominator products \\
\addlinespace[2pt]
150 & Conjunct used as guard & Biconditionals instead of implications, unsatisfiable when x equals y and n prime \\
\addlinespace[2pt]
151 & Type-system mismatch & Rat mod makes isEven always true, skip branch drops tail sum \\
\addlinespace[2pt]
152 & Conjunct used as guard & Equal-length input assumption asserted as conjunct not precondition guard \\
\addlinespace[2pt]
153 & Nat arithmetic truncation & Nat subtraction saturates negative extension strength to zero \\
\addlinespace[2pt]
154 & Precedence bug & Missing parens make \ensuremath{\to} bind inside \ensuremath{\leftrightarrow}, unsatisfiable for empty b \\
\addlinespace[2pt]
155 & Universal over invalid domain & Missing num = 0 base case makes even-digit recurrence unsatisfiable \\
\addlinespace[2pt]
156 & Extra/over-restrictive constraint & Lowercase check in antecedent makes spec vacuously satisfiable \\
\addlinespace[2pt]
158 & Conjunct used as guard & Empty-string result biconditional unsatisfiable on empty word input \\
\addlinespace[2pt]
159 & Missing constraint & Nat subtraction lets number=0 under-constrain result[0] \\
\addlinespace[2pt]
161 & Vacuous disjunct & Disjunction of case implications is vacuous, needs \ensuremath{\land} \\
\addlinespace[2pt]
163 & Conjunct used as guard & Length bound as top-level conjunct forbids empty branch \\
\addlinespace[2pt]
\end{longtable}

\begin{longtable}{@{}>{\raggedright\arraybackslash}p{0.30in}>{\raggedright\arraybackslash}p{1.45in}>{\raggedright\arraybackslash}p{3.50in}@{}}
\caption{Specification \flagMISMATCH{} findings (13 entries).}\label{tab:all-mismatches}\\
\toprule
\textbf{ID} & \textbf{Taxonomy} & \textbf{Summary} \\
\midrule
\endfirsthead
\caption[]{Specification \flagMISMATCH{} findings (13 entries). (continued)}\\
\toprule
\textbf{ID} & \textbf{Taxonomy} & \textbf{Summary} \\
\midrule
\endhead
\midrule
\multicolumn{3}{r}{\emph{Continued on next page}}\\
\endfoot
\bottomrule
\endlastfoot
0 & Edge-case behavior & GT accepts any result on short lists, generated forces false \\
\addlinespace[2pt]
1 & Precondition gap\newline Structural-op choice & Disagreement over non-paren non-space characters in input \\
\addlinespace[2pt]
6 & Structural-op choice\newline Precondition gap & GT keeps empty split tokens and leaves deep prefixes unconstrained \\
\addlinespace[2pt]
17 & Structural-op choice\newline Precondition gap & GT counts empty slots and forces off-legend tokens to zero \\
\addlinespace[2pt]
19 & Structural-op choice\newline Precondition gap\newline Ordering vs multiset & Whitespace tokenizing plus multiset versus membership equality disagreement \\
\addlinespace[2pt]
21 & Edge-case behavior & Divergence on all-equal lists where min equals max \\
\addlinespace[2pt]
29 & Ordering vs multiset & GT multiset reading, generated requires order preservation \\
\addlinespace[2pt]
30 & Ordering vs multiset & GT ignores order, generated requires input order preserved \\
\addlinespace[2pt]
32 & Algorithmic variant & GT allows approximate root, generated requires exact zero \\
\addlinespace[2pt]
57 & Edge-case behavior\newline Algorithmic variant & Trivial-list guard and strict vs non-strict monotonicity on plateaus \\
\addlinespace[2pt]
87 & Ordering vs multiset & GT allows duplicate coordinates, generated forbids via strict order and Nodup \\
\addlinespace[2pt]
120 & Precondition gap & Bounds as preconditions vs constraints impl must uphold \\
\addlinespace[2pt]
157 & Precondition gap & GT rejects degenerate triples, generated specs guard or accept them \\
\addlinespace[2pt]
\end{longtable}

\begin{longtable}{@{}>{\raggedright\arraybackslash}p{0.30in}>{\raggedright\arraybackslash}p{1.45in}>{\raggedright\arraybackslash}p{3.50in}@{}}
\caption{Fixed benchmark-base \flagISSUE{} findings (4 entries).}\label{tab:all-base-issues}\\
\toprule
\textbf{ID} & \textbf{Taxonomy} & \textbf{Summary} \\
\midrule
\endfirsthead
\caption[]{Fixed benchmark-base \flagISSUE{} findings (4 entries). (continued)}\\
\toprule
\textbf{ID} & \textbf{Taxonomy} & \textbf{Summary} \\
\midrule
\endhead
\midrule
\multicolumn{3}{r}{\emph{Continued on next page}}\\
\endfoot
\bottomrule
\endlastfoot
32 & Missing input precondition & omits rational-root existence precondition \\
\addlinespace[2pt]
39 & Open mathematical problem & correctness \ensuremath{\forall}n needs open prime-Fibonacci conjecture \\
\addlinespace[2pt]
123 & Open mathematical problem & spec ties Collatz-orbit completeness to the open Collatz conjecture; theorem unprovable \\
\addlinespace[2pt]
160 & Fixed helper defect & applyOp returns none on division by zero, unsatisfiable \\
\addlinespace[2pt]
\end{longtable}

\endgroup

\end{document}